\documentclass{article}
\usepackage{arxiv}

\usepackage{amssymb,amsmath,latexsym,amsfonts,amsthm}
\usepackage{mathtools}
\usepackage[utf8]{inputenc} 
\usepackage[T1]{fontenc}    

\usepackage{subcaption}
\usepackage{float}
\usepackage{xcolor}
\usepackage{multirow}
\usepackage{graphicx}
\usepackage{hyperref}
\usepackage{mathrsfs}  

\usepackage[linesnumbered,ruled,vlined]{algorithm2e}
\usepackage{float}
\usepackage{upgreek}

\usepackage{multicol}

\usepackage{caption}
\usepackage{subcaption}
\usepackage[utf8]{inputenc}

\title{Closeness and Uncertainty Aware Adversarial Examples Detection in Adversarial Machine Learning}

\author{
Omer Faruk Tuna \\
  Isik University\\
  Istanbul, Turkey \\
  \texttt{omer.tuna@isikun.edu.tr} \\
  \And
 Ferhat Ozgur Catak \\
  Simula Research Laboratory\\
  Fornebu, Norway \\
  \texttt{ozgur@simula.no} \\
  \And
 M. Taner Eskil \\
  Isik University\\
  Istanbul, Turkey \\
  \texttt{taner.eskil@isikun.edu.tr} 
}

\usepackage{graphicx}
\usepackage{multirow}
\usepackage{caption}
\usepackage{subcaption}
\usepackage{mathtools}
\usepackage{amsfonts}

\begin{document}

%
%

\maketitle
\begin{abstract}

While state-of-the-art Deep Neural Network (DNN) models are considered to be robust to random perturbations, it was shown that these architectures are highly vulnerable to deliberately crafted perturbations, albeit being quasi-imperceptible. These vulnerabilities make it challenging to deploy DNN models in security-critical areas. In recent years, many research studies have been conducted to develop new attack methods and come up with new defense techniques that enable more robust and reliable models. In this work, we explore and assess the usage of different type of metrics for detecting adversarial samples. We first leverage the usage of moment-based predictive uncertainty estimates of a DNN classifier obtained using Monte-Carlo Dropout Sampling. And we also introduce a new method that operates in the subspace of deep features extracted by the model. We verified the effectiveness of our approach on a range of standard datasets like MNIST (Digit), MNIST (Fashion) and CIFAR-10. Our experiments show that these two different approaches complement each other, and the combined usage of all the proposed metrics yields up to 99 \% ROC-AUC scores regardless of the attack algorithm.
\end{abstract}

\section{Introduction}
\label{introduction}

Machine learning (ML) applications are transforming our everyday lives, and artificial intelligence (AI) technology is becoming an integral part of our civilization. As AI technology advances, it becomes a key component of many sophisticated tasks that directly affect humans. In the last few years, deep neural networks (DNNs) achieved state-of-the-art performances on the different number of supervised learning tasks, which led them to become widely used in many fields such as medical diagnosis, computer vision, machine translation, speech recognition and autonomous vehicles \cite{lecun2015deeplearning,LE2020417,SHARMA2020100301,AKCAY202056}. However, there are severe concerns about making DNN architectures an integral part of our lives while ensuring the utmost security and reliability requirements. 

Although DNN's have proven their usefulness in real-world applications for many complex problems, they have thus far failed to overcome the challenges faced by deliberately manipulated data, which are known as adversarial inputs \cite{8718038}. Szegedy et al. \cite{szegedy2014intriguing} were among the first who observed the presence of adversarial examples in the image classification domain. The authors have shown that it is possible to perturb an image by a tiny amount to change the decision of the DNN model. It turns out that a very small and quasi-imperceptible perturbation of the input is sufficient to fool the most advanced classifiers and result in the wrong classification. Back then, many studies have been pursued in this new research field named "Adversarial Machine Learning", and these studies were not only limited to the image classification domain. To give some example, in the NLP domain, Sato et al. \cite{sato2018interpretable} showed that it is possible to fool a sentiment analysis model which is trained on textual data by just changing only one word from the input sentence. Another example is in the audio domain \cite{carlini2018audio}, in which the authors constructed targeted adversarial audio samples in automatic speech recognition task by adding very small perturbation to the original waveform. This study demonstrated that the target model could easily be manipulated to transcribe the input as any chosen phrase.

Therefore, a recent problem facing the ML community is to furnish the state of the art algorithms with tools that actively detect and avert adversarial attacks, making them robust to such inputs \cite{WANG2020102634}. Attacks utilizing the vulnerability of DNNs can seriously hamper the security of ML-based systems, sometimes with devastating consequences. In the case of medical applications, a malicious attack can lead to an incorrect diagnosis of disease. It thus can cause severe harm to a patient's health and also damage the healthcare economy \cite{finlayson2019adversarial}. In another domain, many state-of-the-art autonomous navigation algorithms use DNNs to drive vehicles in traffic without human intervention while avoiding accidents. A wrong decision by an autonomous navigation algorithm due to an adversarial attack could cause a fatal accident \cite{sitawarin2018darts,morgulis2019fooling}. For this reason, defending against adversarial attempts and increasing the robustness of the DNN architectures without compromising performance is of crucial importance. 

In this study, we aim to analyze different metrics for adversarial sample detection. Our first step is to analyze moment-based predictive uncertainty estimates of a DNN classifier obtained using Monte-Carlo Dropout Sampling. We have investigated various uncertainty metrics such as \textit{Epistemic Uncertainty}, \textit{Aleatoric Uncertainty}, \textit{Scibilic Uncertainty} and \textit{Entropy}. We showed that the quantified uncertainty in prediction time is strongly correlated with the strength of adversarial attempt, but only within a \textit{low confidence window} in which the impact of applied perturbation starts to change the classifier's prediction. Moreover, we observed that for the \textit{high confidence window}, where applied perturbation has limited impact on the uncertainty metrics and predicted class, another metric based on the closeness of low dimensional representation of input samples to their predicted class' representative distribution does a better job in detecting adversarial examples. We showed that the best detection performance is achieved when all the metrics are used together. An empirical validation was systematically conducted in the prediction time for some well-known adversarial machine learning attacks on standard datasets.

To sum up; our main contributions for this paper are:

\begin{itemize}

	\item To the best of our knowledge, we are the first in the research community who investigated the use of aleatoric and scibilic uncertainty for the purpose of detecting adversarial samples.
	
	\item We introduce a novel method to quantify the closeness of an input sample's representation with its predicted class data distribution in the subspace of last hidden layer activations. 
	\item We experimentally show that there is no single metrics that works well in every condition, and an ensemble approach of using different metrics should be considered.
\end{itemize}

The rest of the paper is organized as follows: Section \ref{ch:related_work} will introduce some of the known attack types and defense techniques in the literature. In Section \ref{ch:preliminaries}, we will introduce the notion of uncertainty together with its main types and discuss how we can quantify different uncertainty metrics for a DNN classifier. Section \ref{ch:approach} will give the details of our approach. We will present our experimental results in Section \ref{ch:results} and conclude our work in Section \ref{ch:conclusion}.
Codes for this study are released on GitHub \footnote{\url{https://github.com/omerfaruktuna/adversarial-detection}} for scientific use.

\section{Related Work}
\label{ch:related_work}

\subsection{Adversarial attacks}

\begin{figure}[!htbp]
    \centering
    \includegraphics[width=0.8\linewidth]{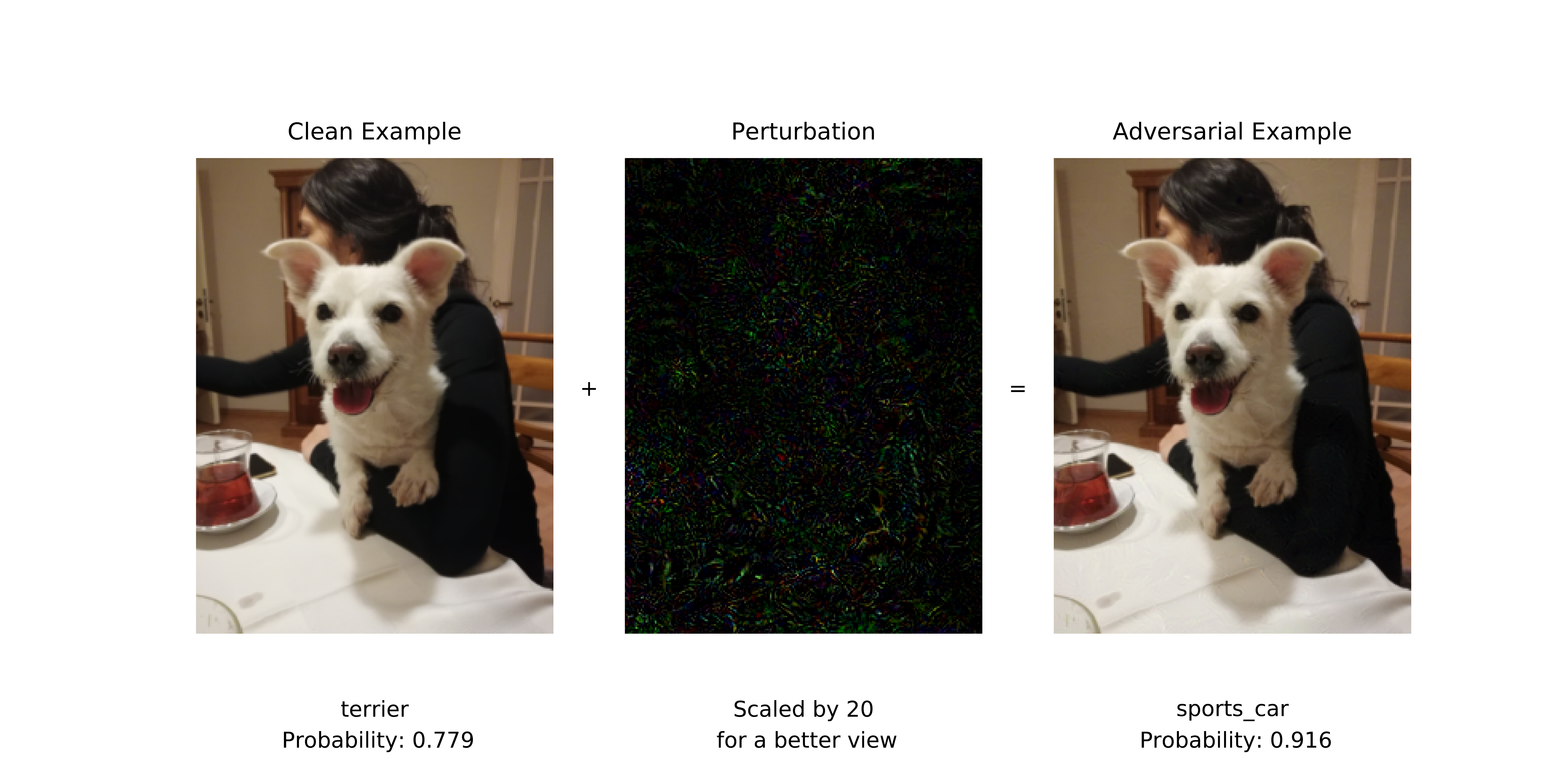}
    \caption{The figure shows an adversarial attack example. The malicious perturbation is applied upon the original image. The precisely crafted perturbation manipulates the model in such a way that a "Terrier (Dog)" is wrongly classified as "Sports Car" with very high degree of confidence.}
    \label{fig:adv-ml-ex}
\end{figure}

Deep learning models contain many vulnerabilities and weaknesses, making them difficult to defend in adversarial machine learning. For instance, they are often sensitive to small changes in the input data, resulting in unexpected results in the model's final output. Figure \ref{fig:adv-ml-ex} shows how an adversary would exploit such a vulnerability and manipulate the model through the use of carefully crafted perturbation applied to the input data.

The attack strategies are mainly based on perturbing the input instance to maximize the model's loss.  Many adversarial attack algorithms have been proposed in the literature in the last few years. The well-known adversarial attacks used in this study are \texttt{Fast-Gradient Sign Method, Iterative Gradient Sign Method, Projected Gradient Descent, Carlini{\&}Wagner}, and \texttt{DeepFool}. Section \ref{sec:fgsm-definition} - \ref{sec:deepfool-definition} briefly describes these five adversarial machine learning attacks.

\subsubsection{Fast-Gradient Sign Method}\label{sec:fgsm-definition}
This method, also known as FGSM \cite{goodfellow2015explaining}, is one of the earliest and most popular adversarial attacks to date. FGSM utilizes the derivative of the model's loss function for the input image to determine in which direction the pixel values of the input image should be altered to minimize the loss function of the model. Once this direction is extracted, it changes all pixels simultaneously in the opposite direction to maximize the loss. For a model with classification loss function described as $L(\theta,\mathbf{x},y)$ where $\theta$ represents the parameters of the model, $\mathbf{x}$ is the benign input to the model (sample input image in our case), $y_{true}$ is the actual label of our input, we can generate adversarial samples using the formula below:

\begin{equation}
    \mathbf{x}^* = \mathbf{x} + \epsilon \cdot sign\left(\nabla_x L(\theta,\mathbf{x},y_{true}) \right)
\end{equation}

One last key point about FGSM is that it is not designed to be optimal but fast. That means it is not designed to produce the minimum required adversarial perturbation. Besides, this method's success ratio is relatively low in small $\epsilon$ values compared to other attack types.

\subsubsection{Iterative Gradient Sign Method}
Kurakin et al. \cite{kurakin2017adversarial} proposed a small, but effective improvement to the FGSM. In this approach, rather than taking only one step of size $\epsilon$ in the gradient sign's direction, we take several but smaller steps $\alpha$, and we use the given $\epsilon$ value to clip the result. This attack type is often referred to as Basic Iterative Method (BIM), and it is merely FGSM applied to an input image iteratively. Generating perturbed images under $L_{inf}$ norm for BIM attack is given by Equation \ref{eq:bim}.

\begin{equation}
\begin{aligned}
\mathbf{x}^* & = \mathbf{x} \\
\mathbf{x}_{N+1}^* & = \mathbf{x} + Clip_{x, \epsilon} \{ \alpha \cdot sign \left( \nabla_\mathbf{x} L(\mathbf{x}_N^*, y_{true}) \right) \}
\end{aligned}
\label{eq:bim}
\end{equation}
where $\mathbf{x}$ is the input sample, $\mathbf{x}^*$ is the produced adversarial sample at $i$\textsuperscript{th} iteration, $L$ is the loss function of the model, $y_{true}$ is the actual label for input sample, $\epsilon$ is a tunable parameter, limiting maximum level of perturbation in given $l_{inf}$ norm, and $\alpha$ is the step size. 

The success ratio of BIM attack is higher than the FGSM  \cite{DBLP:journals/corr/KurakinGB16a}. By adjusting the $\epsilon$ parameter, the attacker can have a chance to manipulate how far an adversarial sample will be pushed past the decision boundary. 

One can group BIM attacks under two main types, namely BIM-A and BIM-B. In the former type, we stop iterations as soon as we succeed in fooling the model (passing the decision boundary). In the latter, we continue the attack until the end of the provided number of iterations to push the input further away from the decision boundary. Predominantly the second option is preferred by the attacker to produce more confidently wrong predictions, as we implemented in this study.

\subsubsection{Projected Gradient Descent}
This method, also known as PGD, has been introduced by Madry et al. \cite{madry2019deep}. It perturbs a clean image $\mathbf{x}$ for several numbers of $i$ iterations with a small step size in the direction of the model's loss function's gradient. Different from BIM, after each perturbation step, it projects the resulting adversarial sample back onto the $\epsilon$-ball of input sample, instead of clipping. Moreover, instead of starting from the original point ($\epsilon$=0, in all dimensions), PGD uses random start, which can be described as:  

\begin{equation}
    \mathbf{x}_0 = \mathbf{x} + U\left( -\epsilon, +\epsilon \right)
\end{equation}
where $U\left( -\epsilon, +\epsilon \right)$ is the uniform distribution between ($-\epsilon, +\epsilon$).



\subsubsection{Carlini {\&} Wagner Attack}

This attack type has been introduced by Carlini and Wagner \cite{carlini2017evaluating}, and it is one of the most powerful attack types to date. Therefore, it is generally used as a benchmark for the adversarial defense research community that aims to create more robust DNN architectures resistant to adversarial attempts. CW attack achieves higher success rates on typically trained models for most well-known datasets. It can fool defensively distilled models as well, on which other attack types barely succeed in crafting adversarial examples.

The authors redefine the adversarial attack as an optimization problem that can be solved using gradient descent to craft more powerful and effective adversarial samples under different $L_{p}$ norms. 

\subsubsection{Deepfool Attack} \label{sec:deepfool-definition}

This attack type has been proposed by Moosavi-Dezfooli et al. \cite{moosavidezfooli2016deepfool} and it is one of the powerful attack types in literature. It is designed to be used in different distance norm metrics such as $L_{inf}$ and $L_{2}$ norms. 

Deepfool attack has been designed based on the assumption that neural network models behave as a linear classifier and the classes are separated by a hyperplane. The algorithm starts from the initial input point $\mathbf{x_t}$ and at each iteration, it calculates the closest hyperplane and the minimum perturbation amount, which is the orthogonal projection to the hyperplane. Then the algorithm calculates $\mathbf{x}_{t+1}$ by adding the minimal perturbation to the $\mathbf{x}_{t}$  and checks whether misclassification is achieved.

Adversarial machine learning is a highly active research area, and we see new adversarial attack algorithms are being proposed intensely, which we couldn't mention here. Some recent studies are Square Attack \cite{andriushchenko2020square}, HSJA \cite{9152788}, Bandit \cite{ilyas2019prior}. Besides, some recent studies utilize MC Dropout sampling and uncertainty information to craft adversarial samples. Liu et al. \cite{9008259} proposed Universal Adversarial Perturbation (UAP) method that utilizes a metric called virtual Epistemic uncertainty obtained from the model's structural activation. However, estimating the model's uncertainty involves aggregating all the neurons' virtual Epistemic uncertainties, which is computationally costly. And finally, Tuna et al. \cite{tuna2021exploiting} proposed several iterative attack variants based on the model's quantified epistemic uncertainty obtained from the model's final softmax scores.

\begin{figure}[!htbp]
 \centering
	\includegraphics[width=1.0\linewidth]{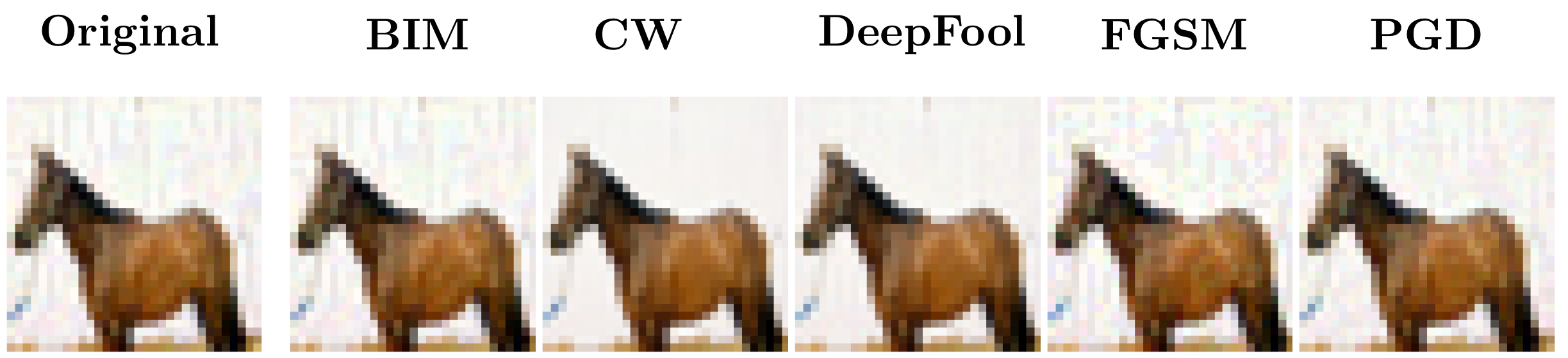}
	\caption{Samples.}
	\label{fig:dummy_plot}
\end{figure}

Figure \ref{fig:dummy_plot} shows some of the adversarial examples crafted by attack algorithms explained above.

\subsection{Adversarial defense}

Since the discovery of DNN's vulnerability to adversarial attacks \cite{szegedy2014intriguing}, a vast amount of research has been conducted on defending against these attacks. Defence against adversarial attacks can be divided into two categories; (i) improving the robustness of classifiers to existing attack types, (ii) methods for detecting adversarial samples. Inline with the aim of this study, we focus on the detection of the adversarial samples and below, we briefly mention some of the notable adversarial example detection approaches proposed recently.

Feinman et al. \cite{feinman2017detecting}  focused on detecting adversarial samples through two measures; estimations of epistemic uncertainty and the kernel density. They formulated the uncertainty estimate as the variance of a Bayesian distribution which is obtained from the neural network model with dropout. They estimated the kernel density through the activations of the last hidden layer.  However, no other uncertainty metrics have been analyzed apart from epistemic uncertainty and tuning the bandwidth for kernel density estimation method is a serious issue. Ma et al. \cite{ma2018characterizing}  proposed to detect adversarial samples using an auxiliary classifier that is trained to use an expansion-based measure; local intrinsic dimensionality. Metzen et al. \cite{metzen2017detecting} proposed augmenting a DNN with an additional detector subnetwork, trained on the binary classification task of normal and adversarial samples. Yang et al. proposed a framework called ML-Loo \cite{yang2019mlloo} for detecting adversarial examples through thresholding a scale estimate of feature attribution scores from Leave-One-Out (LOO).  Lee et al. proposed a method \cite{lee2018simple} for detecting both out-of-distribution samples and adversarial attacks.
Meng et al. \cite{meng2017magnet} proposed a defense method that comprises two components: detector and reformer. The former is used to inspect input samples and determine if they are benign or not, and the latter is used to take inputs classified as benign by the detector and reform them to remove any remaining adversarial nature. Although the authors show the efficacy of their defense against different adversarial attacks, later it was shown that their defense method is vulnerable to CW attack \cite{carlini2017magnet}. And some other notable studies in literature for detecting adversarial samples are \cite{CarraraFCAB19} and \cite{8482346}.

\section{Preliminaries} 
\label{ch:preliminaries}

We start this section by first introducing the main types of uncertainty in machine learning. And then, we will continue by presenting how the uncertainty metrics can be quantified in the context of deep learning.

\subsection{Uncertainty in Machine Learning}\label{sec:preliminaries}

There are two main forms of uncertainty in machine learning: epistemic uncertainty and aleatoric uncertainty \cite{hullermeier2020aleatoric,AN2020110617,ZHENG2021107046}.   

\subsubsection{Epistemic Uncertainty}

\begin{figure*}[!htbp]
 \centering
	\includegraphics[width=0.8\linewidth]{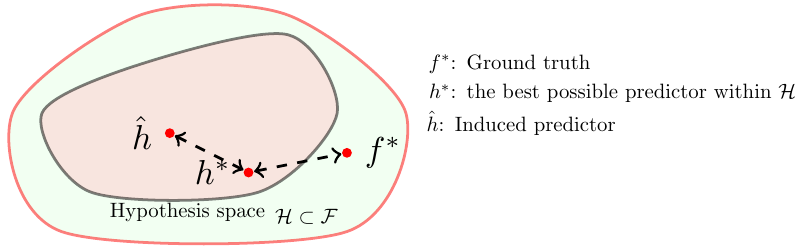}
	\caption{Different types of Epistemic Uncertainty.}
	\label{fig:epistemic-uncertainties}
\end{figure*}

Epistemic uncertainty relates to uncertainty caused by a lack of knowledge and limited data needed for a perfect predictor \cite{ANTONELLI2020746}. It can be categorized under 2 groups as \textit{approximation uncertainty} and \textit{model uncertainty} as depicted in Figure \ref{fig:epistemic-uncertainties}.


\paragraph{Approximation Uncertainty}

In a conventional machine learning task, the learner is given data points from an independent, identically distributed dataset. Then he/she tries to induce a hypothesis $\hat{h}$ from the hypothesis space $\mathcal{H}$ by picking a proper learning method with its related hyperparameters and minimizing the expected loss (risk) with a selected loss function, $\ell$. However, what the learner does is to try to minimize the \textit{empirical risk} ${R}_{emp}$ which is an estimate of real risk $R(h)$. The induced $\hat{h}$ is an approximation of the $h^{*}$ which is the optimum hypothesis within $\mathcal{H}$ and the real risk minimizer. This fact results in an approximation uncertainty.

\paragraph{Model Uncertainty}

Suppose the chosen hypothesis space $\mathcal{H}$ does not include the perfect predictor. In this case, the learner has no chance to realize his/her objective of discovering a hypothesis function that can successfully map all possible inputs to outputs. This drives to an inconsistency between the ground truth $f^{*}$ and the best possible function $h^{*}$ within $\mathcal{H}$, called model uncertainty. 

However, Universal Approximation Theorem states that for any target function $f$, a neural network can approximate $f$  \cite{zhou2018universality,Cybenko}. The hypothesis space $\mathcal{H}$ is huge for deep neural networks. Hence it will not be wrong to assume that $h^{*} = f^{*}$. One can disregard the model uncertainty for deep neural networks, and may only care about the approximation uncertainty. Consequently, in deep learning tasks, the actual source of epistemic uncertainty is linked to approximation uncertainty. Epistemic uncertainty points to the confidence a model has about its prediction \cite{Loquercio_2020}. The underlying cause is the uncertainty about the parameters of the model. This type of uncertainty is apparent in the regions with limited training data, and the model weights are not optimized correctly.

\subsubsection{Aleatoric Uncertainty}

Aleatoric uncertainty refers to the variability in an experiment's outcome, which is due to the inherent random effects \cite{GUREVICH2019291}. This type of uncertainty can not be reduced albeit having enough training samples \cite{SENGE201416}. A perfect example for this phenomenon is the noise observed in the measurements of a sensor. 


\begin{figure}[!htbp]
 \centering
	\includegraphics[width=0.8\linewidth]{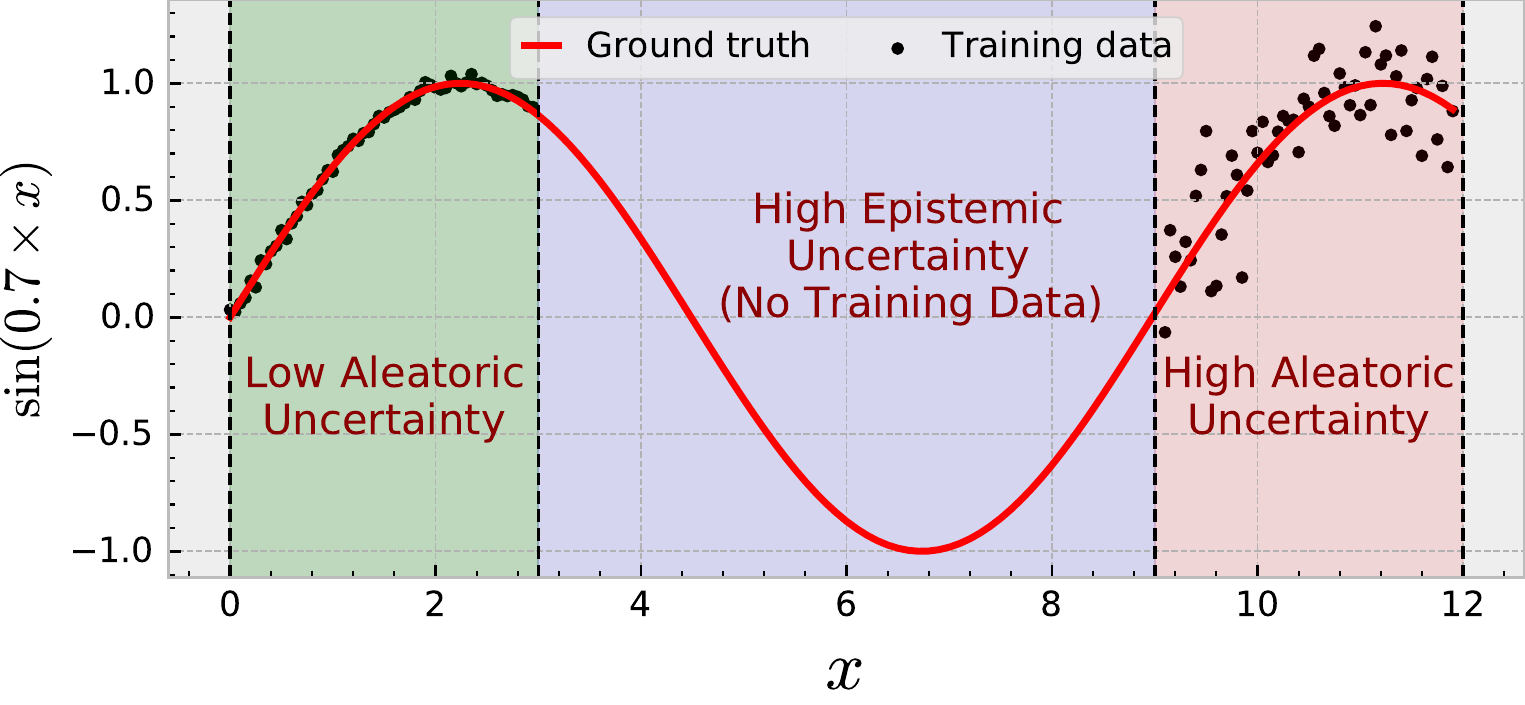} 
	\caption{Illustration of the Epistemic and Aleatoric uncertainty.}
	\label{fig:uncertainties_lat}
\end{figure}

Figure \ref{fig:uncertainties_lat} shows a simple nonlinear function ( $\sin({0.7 \times x})$ where $x\in[0,12]$ ) plot. As shown in the region where data points are populated at right ($9<x<12$), the noisy samples are clustered, leading to high aleatoric uncertainty. For example, these points may represent a faulty sensor measurement; one can conclude that the sensor produces errors around $x=10.5$ for some inherent reason. We can also conclude that the middle regions of the figure represent the high epistemic uncertainty areas. Because there are not enough training samples for our model to describe the data best. Moreover, we can claim that the high epistemic uncertainty area represents the low prediction accuracy area.

\subsubsection{Scibilic Uncertainty}

Reinhold et al. \cite{reinhold2020finding} combined epistemic and aleatoric uncertainty into a new type of uncertainty which they call \textit{scibilic uncertainty}. This metric has been used in an image segmentation task to spot the areas in an input image which the model could figure out how to predict if it was trained with enough data. Once we quantify epistemic and aleatoric uncertainty, we can calculate scibilic uncertainty by dividing the first one by the latter. A DNN model trained on clean (natural occurring) data might result into high epistemic uncertainty for a suspicious input. However, the model can result in high aleatoric uncertainty for that same input due to some intrinsic property of the data and thus face difficulty in making a reliable prediction. Division operation helps us to retain epistemic uncertainty which is not caused by the mentioned difficulty of the model for that specific input.


\subsubsection{Entropy}
As a well-known concept from information technology, the entropy of a random variable is a measure of the average level of randomness or uncertainty inherent in the possible outcomes \cite{shannon1948mathematical}. 

\subsection{Quantifying Uncertainty in Deep Neural Networks}\label{sec:preliminaries}

In recent years, various research studies have been performed to quantify uncertainty in deep learning models. Most of the work was based on Bayesian Neural Networks, which learn the posterior distribution over weights to quantify predictive uncertainty \cite{Hinton1995BayesianLF}. However, the Bayesian NN's come with additional computational cost and inference issue. Therefore, several approximations to Bayesian methods have been developed which make use of variational inference \cite{NIPS2011_7eb3c8be,paisley2012variational,hoffman2013stochastic,blundell2015weight}. On the other hand, Lakshminarayanan et al. \cite{lakshminarayanan2017simple} used the deep ensemble approach as an alternative to Bayesian NN's to quantify predictive uncertainty. But this approach requires training several NN's which may not be feasible in practice. A more efficient and elegant approach was proposed by Gal et al. \cite{gal2016dropout}. The authors showed that a neural network model with inference time dropout is equivalent to a Bayesian approximation of the Gaussian process. 

In \cite{kendall2017uncertainties}, Kendall and Gal proposed an approach to capture both epistemic and aleatoric uncertainties in a single model. They used a bayesian neural network $f$ (CNN architecture) with weights denoted by $\hat{\omega}$ which maps an input $x$ to $\hat{y}$ and ${\sigma}^2$. In the authors' approach, the model output is split into two parts as predictive mean $(\hat{y})$ and predicted variance $\hat{\sigma}^2$ terms and the two types of uncertainty are estimated as below:

\begin{equation}
    \underbrace{\frac{1}{T} \sum_{t=1}^T diag(\hat{\sigma}^2)}_{aleatoric} + \underbrace{\frac{1}{T} \sum_{t=1}^T (\hat{y} - \bar{y})^{\otimes 2}}_{epistemic}
\end{equation}

where $\bar{y}=\sum_{t=1}^T \hat{y}_{t}/T$ and $y^{\otimes 2} = yy^T$

The above approach is exquisite and shown to be effective in computer vision tasks like image segmentation. However, since the model output is split into two parts for predicting mean and variance terms, it was not convenient for us to use in adversarial machine learning experiments. Because, the attack algorithms are designed to work for model architectures with only one output part (only prediction output, no variance part) which led us to seek alternatives. 

In \cite{KWON2020106816}, Kwon et al. proposed an alternative way of capturing both aleatoric and epistemic uncertainty for a classification model. In their approach, the variance of the prediction is composed of two terms representing aleatoric and epistemic uncertainty respectively. Let $\hat{\omega}$ represents parameters (learnt weights) used in the neural network, the number of different output classes is denoted by K, then the prediction $y^*$ of a model for any test sample $x^*$ given the weights of the model is denoted by $p(y^*|x^*,\hat{\omega})$ where $y^*  \in \mathbb{R}^{k}$. The formulation for their method is given below:





\begin{equation}
    Var_{p(y^*|x^*,\omega)}(y^*) = \mathbb{E}_{p(y^*|x^*,\omega)}(y^{*{\otimes 2}}) - \mathbb{E}_{p(y^*|x^*,\omega)}(y^*)^{\otimes 2}
\end{equation}

\begin{equation}
    Var_{p(y^*|x^*,\omega)}(y^*) = diag\{\mathbb{E}_{p(y^*|x^*,\omega)}(y^*)\} - \mathbb{E}_{p(y^*|x^*,\omega)}(y^*)^{\otimes 2}
\end{equation}

\begin{equation}
    = \underbrace{\frac{1}{T} \sum_{t=1}^T [diag\{p(y^*|x^*,\hat{\omega}_t)\} - p(y^*|x^*,\hat{\omega}_t)^{\otimes 2}]}_{aleatoric}
    \label{eq:alea}
\end{equation}

\begin{equation}
    \underbrace{+ \frac{1}{T} \sum_{t=1}^T \{p(y^*|x^*,\hat{\omega}_t)\} - \hat{p}(y^*|x^*,\hat{\omega}_t)^{\otimes 2}}_{epistemic}
    \label{eq:epis}
\end{equation}

where, $\hat{p}(y^*|x^*,\hat{\omega}_t) = \sum_{t=1}^T \{p(y^*|x^*,\hat{\omega}_t)\}$

Both of the above equations (\ref{eq:alea} and \ref{eq:epis}) output a matrix of shape $k \times k$ where the diagonal elements represent the variance of each output class and we used the mean of the diagonal terms for quantifying uncertainty metrics for a given input $\mathbf{x}^*$.

Once the quantification of epistemic and aleatoric uncertainty is over, we calculate scibilic uncertainty as below:

\begin{equation}
   Scibilic = Epistemic \, / \, Aleatoric
\label{eq:sc}
\end{equation}

Finally, we compute entropy as below:

\begin{equation} 
   H(p_T(\mathbf{x}^*)) = - \sum_{k \in K} p_T(\mathbf{x}^*)[k] \log{(p_T(\mathbf{x}^*)[k])}
   \label{eq:ent}
\end{equation}

where $p_T (\mathbf{x}^*)$  is the average prediction score of $T$ different predictions when dropout is enabled. That is: 

\begin{equation} \label{eq6}
   p_T (\mathbf{x}^*)  = \frac{1}{T} \sum_{t=1}^T \{p(y^*|x^*,\hat{\omega}_t)\}
\end{equation}

This way, instead of calculating entropy over the single softmax prediction output of the model, we consider $T$ different predictions to have a reliable uncertainty (entropy) estimation.

\section{Approach}\label{ch:approach}

For the quantification of uncertainty metrics we used Equation \ref{eq:alea} for aleatoric uncertainty, Equation \ref{eq:epis} for Epistemic Uncertainty, Equation \ref{eq:sc} for Scibilic Uncertainty and Equation \ref{eq:ent} for Entropy. 

Apart from the uncertainty metrics, one other possible way to understand the underlying mechanism of adversarial machine learning vulnerabilities, is to look at the manifold (low dimensional areas where the input data distribution is found) of the data used in the model training phase. High dimensional data like images are known to lie on low dimensional data manifold \cite{10.5555/1557216}. And the manifolds of source classes which are the representations of the input instance in lower-dimensional space become more linear and easy to work with as we go to the deep layers of DNN's \cite{bengio2012better}. For this reason, we opted to work in the feature space of the last hidden layer activations. We used a tricky and trivial approach to understand the closeness of an input instance to the manifold which the predicted class of the input is represented.  We grouped all the last hidden layer output vectors of clean, noisy and perturbed images of each class together and tried to train a secondary model to teach that that all these lower-dimensional representations of input instances correspond to the same class of input. By doing so, we could let our secondary model learn that the representation of any perturbed image in lower-dimensional space should virtually be close to its original class manifold rather than its wrongly predicted class manifold. The details of our approach is as follows:


 Let  $\mathcal{D} = \{(\mathbf{x},y) | \mathbf{x}  \in \mathbb{R}^{m\times n}, \mathbf{y}  \in \mathbb{R}^m\} $ be the training set for our CNN classifier $(H_{cnn})$ consisting of all the clean samples $\{\mathbf{x}_{1},\mathbf{x}_{2},\cdots,\mathbf{x}_{m}\}$, and their corresponding actual labels $\{\mathbf{y}_{1},\mathbf{y}_{2},\cdots,\mathbf{y}_{m}\}$. 
 
 We first apply a noise with normal distribution to all the samples in training set as below:
 
  \begin{equation}
    \eta = \mathcal{N}(0,\epsilon)
\end{equation}

 \begin{equation}
    x\textsuperscript{(noisy)} = x + \eta , \,\,\,\,\, \eta  \in \mathbb{R}^{m\times n}
\end{equation}

We then apply adversarial attack $\mathcal{F}$ (we used BIM) to all the training samples in our original training set with the same $\epsilon$ value that we used in crafting noisy samples. And we get perturbed samples as in below where $\delta$ is the perturbation amount derived from the attack algorithm:
 
   \begin{equation}
    \delta = \mathcal{F}(\mathbf{x})
\end{equation}

  \begin{equation}
    x\textsuperscript{(pert)} = x + \delta , \,\,\,\,\, \delta  \in \mathbb{R}^{m\times n}
\end{equation}
 
After we increase the number of our training samples with crafted noisy and perturbed samples, we feed all these samples to our CNN classifier $H_{cnn}$ to get their corresponding last hidden layer activation outputs $v$:

\begin{equation}
    v\textsuperscript{(clean)} = H_{cnn}(\mathbf{x}\textsuperscript{(clean)})
\end{equation}

\begin{equation}
    v\textsuperscript{(noisy)} = H_{cnn}(\mathbf{x}\textsuperscript{(noisy)})
\end{equation}

\begin{equation}
    v\textsuperscript{(pert)} = H_{cnn}(\mathbf{x}\textsuperscript{(pert)})
\end{equation}

Then, we combine all the last hidden layer activation outputs in one pool to get $\mathbf{V}  \in \mathbb{R}^{3 \cdot m \times j}$as in below, where $j$ is the dimension of the last hidden layer for the CNN model:

\begin{equation}
    \mathcal{V} = v\textsuperscript{(clean)} \cup v\textsuperscript{(noisy)} \cup v\textsuperscript{(pert)} 
\end{equation}

The corresponding labels for these v\textsuperscript{(clean)}, v\textsuperscript{(noisy)} and v\textsuperscript{(pert)} vectors will all be y\textsuperscript{(clean)}. Because we would like to teach our MLP model that all these vectors represent source class' distribution in the sub-space of last hidden layer activations. Therefore, they correspond to actual class labels of the inputs $x$ from which they are derived.  Thus, we just concatenate y\textsuperscript{(clean)} vector multiple times with itself to get $\mathbf{Y}  \in \mathbb{R}^{3 \times m}$

\begin{equation}
    \mathcal{Y} = y\textsuperscript{(clean)} + y\textsuperscript{(clean)} + y\textsuperscript{(clean)} 
\end{equation}

Finally, using the new training set $(V,Y)$ obtained from the last hidden layer activation outputs of clean, noisy and perturbed samples, we train our MLP model $H_{mlp} : \mathcal{V} \mapsto \mathcal{Y}$.

When the training of the MLP model is over, for any test image $x_{test}$, we can get the last hidden activations $v_{test}$ of the CNN model and then feed it to our MLP model to get the softmax score outputs. Softmax score output vector ($\mathcal{O}$) of the MLP model will be of shape $k$, where k is the number of classes for our original training data. And we use the value at the index of the predicted label ($pred$) for the CNN model which is $\mathcal{O}[pred]$ as our last metric value for detecting adversarial samples.

\subsection{Explanatory Research on Uncertainty Quantification Methods}

\begin{figure}[!htbp]
    \centering
    \begin{subfigure}[b]{0.4\linewidth}
         \centering
         \captionsetup{justification=centering}
         \includegraphics[width=1\linewidth]{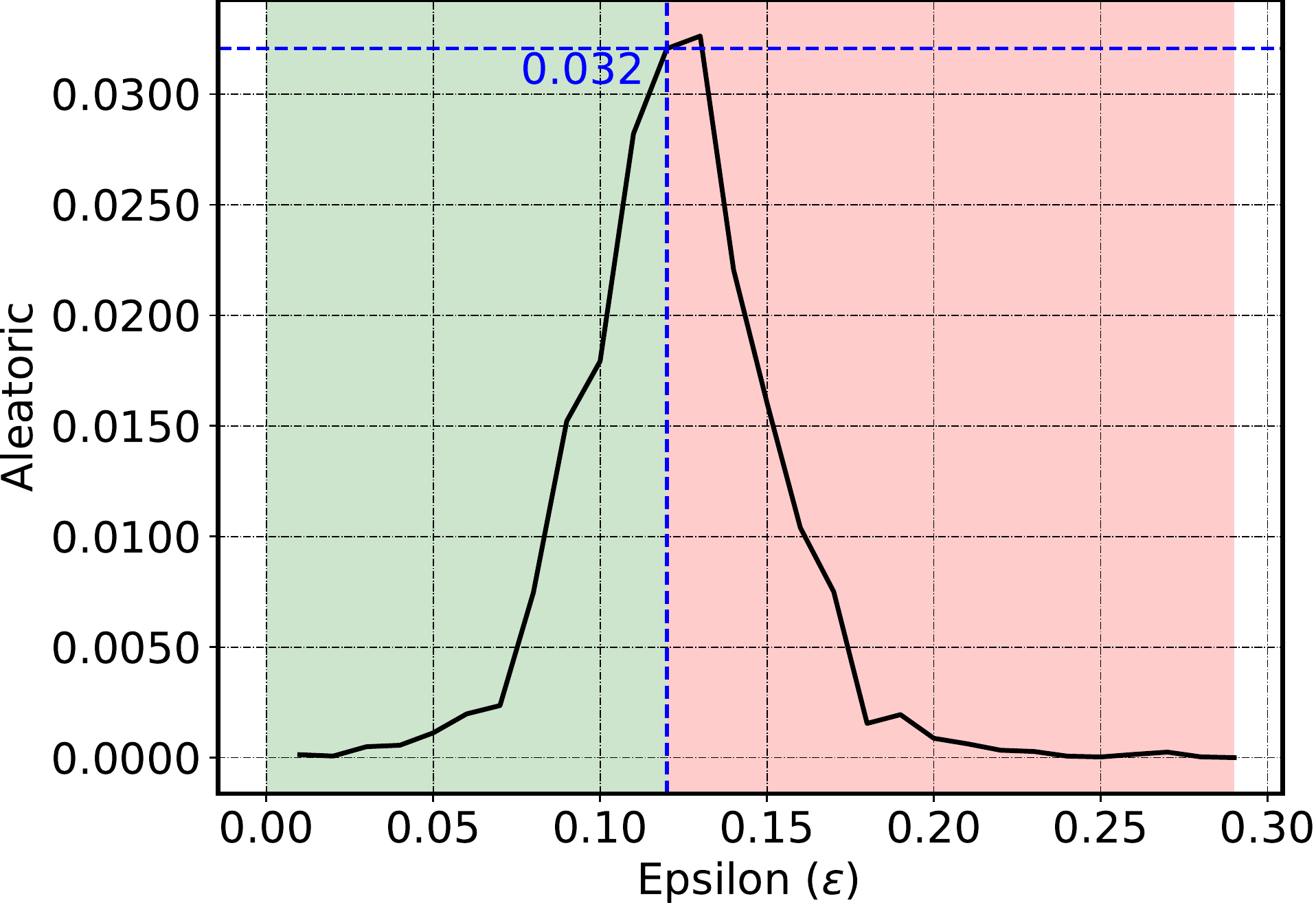}
         \caption{Aleatoric}
	    \label{fig:Aleatoric}
     \end{subfigure}
     \begin{subfigure}[b]{0.4\linewidth}
         \centering
         \captionsetup{justification=centering}
         \includegraphics[width=1\linewidth]{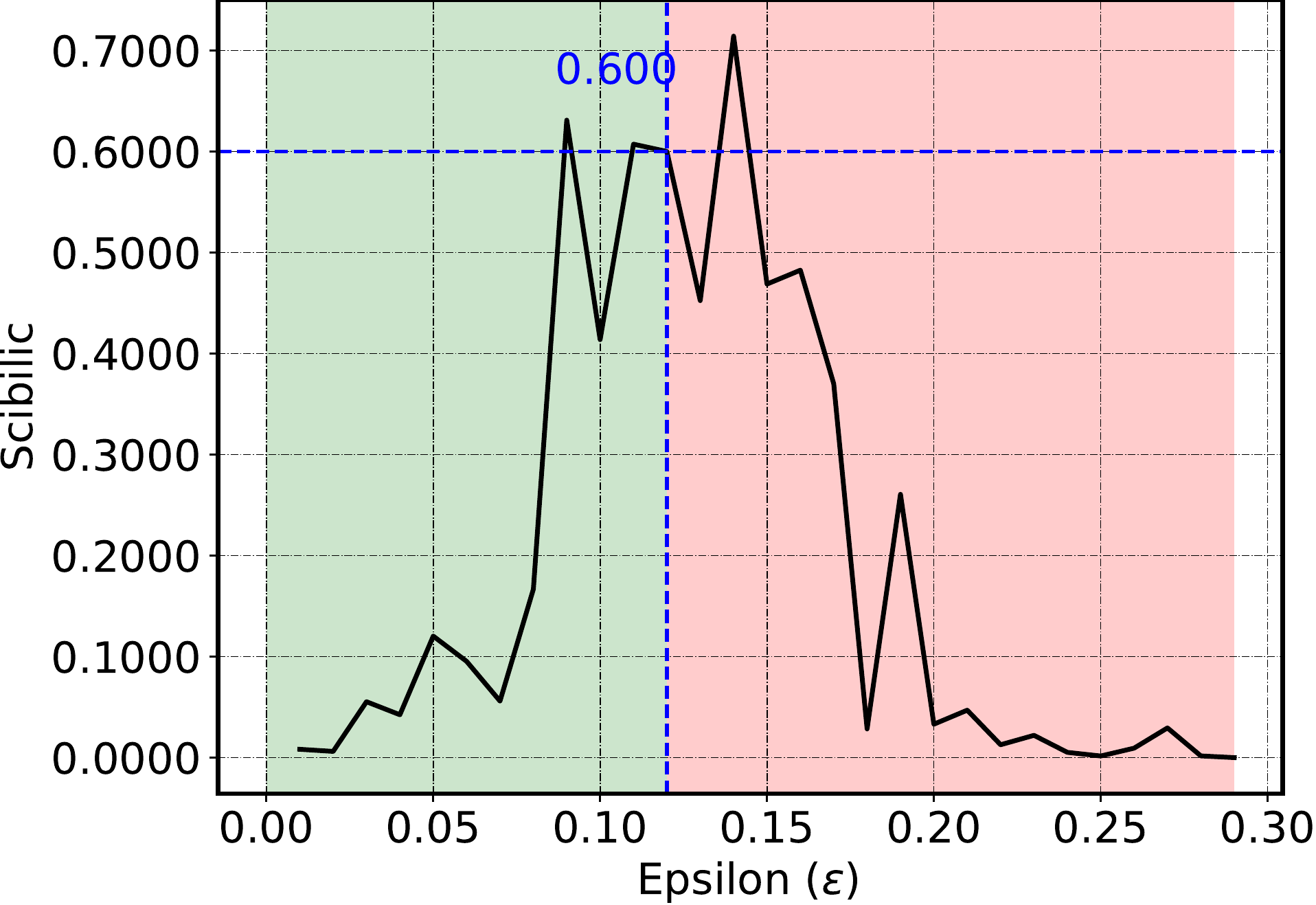}
         \caption{Scibilic}
	    \label{fig:Scibilic}
     \end{subfigure}
     \begin{subfigure}[b]{0.4\linewidth}
         \centering
         \captionsetup{justification=centering}
         \includegraphics[width=1\linewidth]{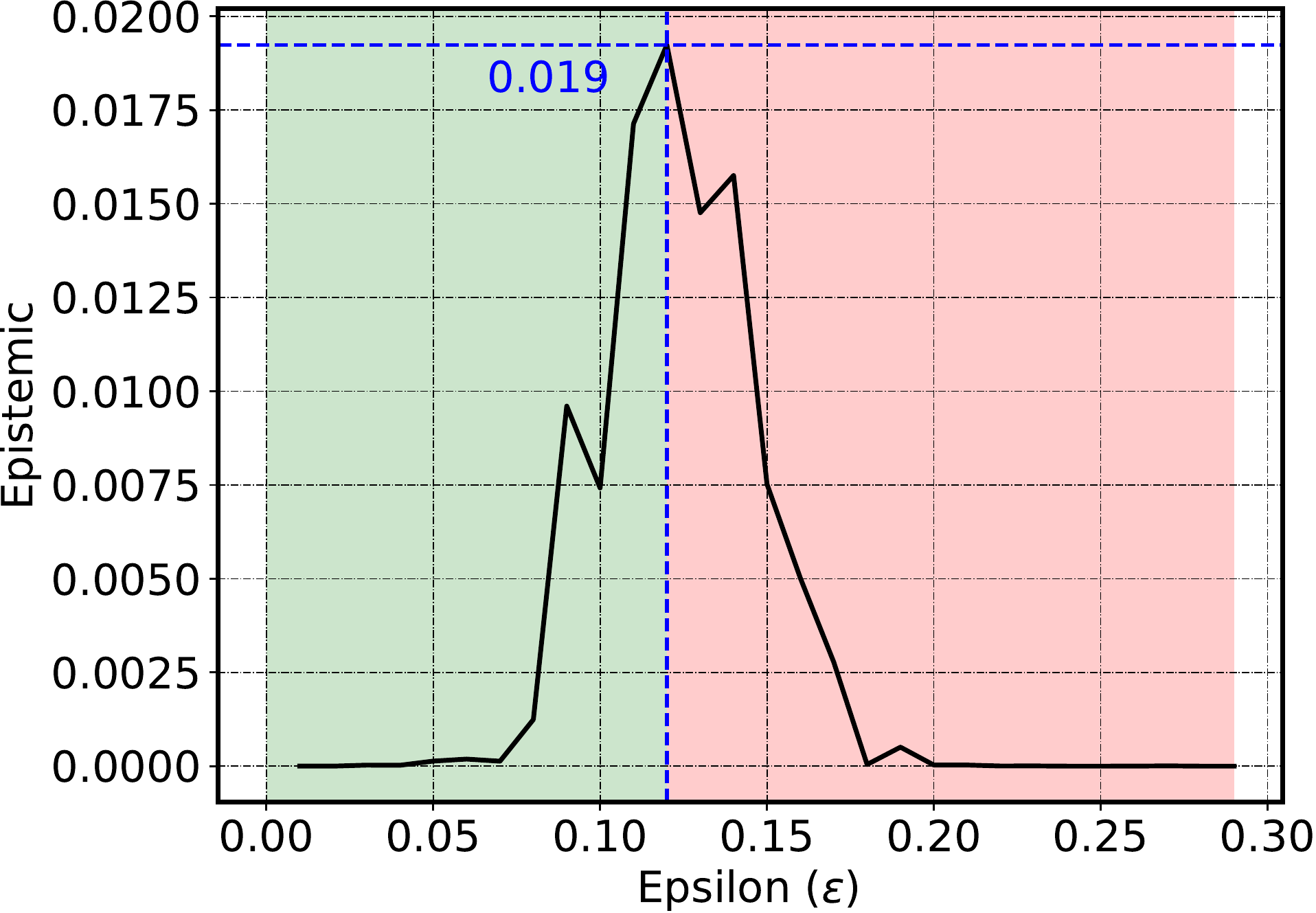}
         \caption{Epistemic}
	    \label{fig:Epistemic}
     \end{subfigure}
     \begin{subfigure}[b]{0.4\linewidth}
         \centering
         \captionsetup{justification=centering}
         \includegraphics[width=1\linewidth]{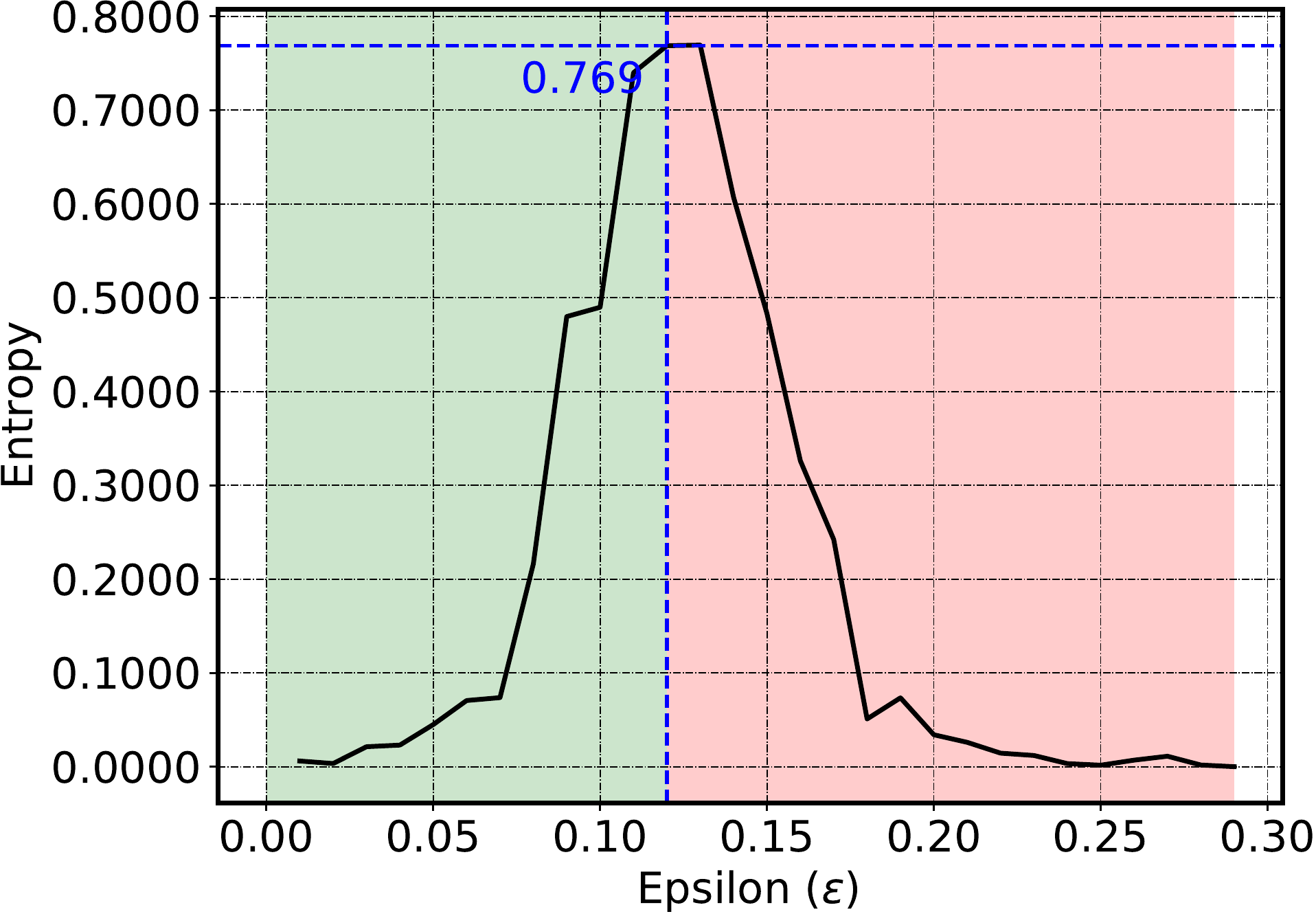}
         \caption{Entropy}
	    \label{fig:Entropy}
     \end{subfigure}
	\caption{Change of uncertanty-based metrics under BIM attack with different amount of maximum allowed perturbation ($\epsilon$) values.}
	\label{fig:uncertainty_values}
\end{figure}

We have made a simple test on a sample image to visualize how the uncertainty metrics behave under an adversarial attack with varying strengths. Figure \ref{fig:uncertainty_values} shows how the uncertainty quantification values affected under BIM attack to our model with different allowed perturbation amounts, $\epsilon$.
We can see that all the quantified uncertainty indicators increases as the amount of perturbation applied to the image become high enough to fool the classifier. Indeed, almost maximum epistemic uncertainty, aleatoric uncertainty and entropy values are observed at an $\epsilon$ value where the model starts mispredicting the input class. We can name this interval as "low confidence interval". Moreover, when the amount of perturbation used to fool the model is high, the model starts to predict the wrong class even more confidently, resulting in a decrease in uncertainty estimates and thus, they are not so reliable for detection as these metrics can not act as a separator anymore. We can call this interval as "high confidence interval". For those cases, we need an additional indicator to help us to increase detection accuracy scores. To overcome this problem, we used another metric which we call $\textit{closeness score for predicted class}$ obtained from the last hidden layer activation's.

\subsection{Explanatory Research on our Closeness Metric}

\begin{figure}[!htbp]
    \centering
    \includegraphics[width=0.6\linewidth]{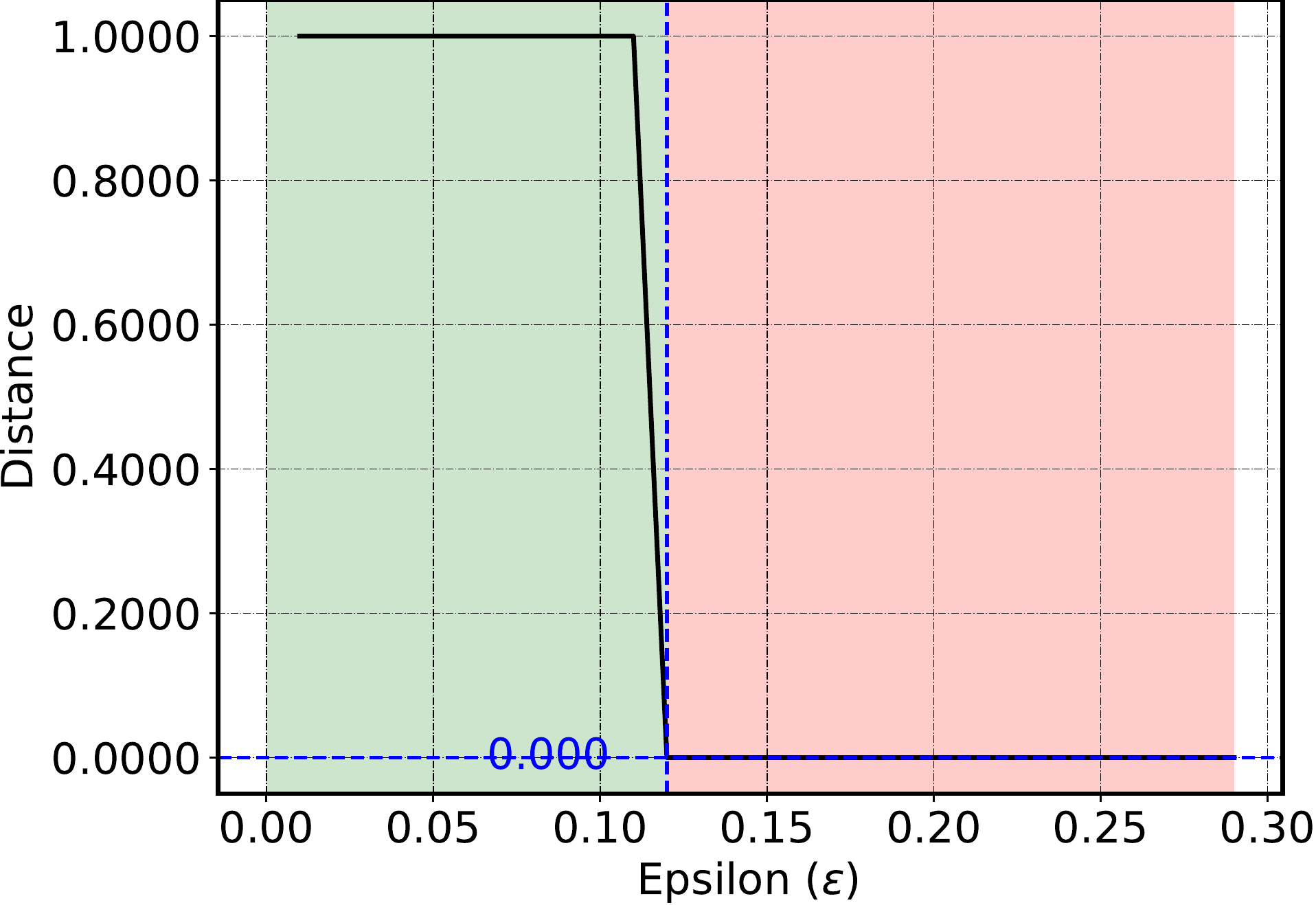}
	\caption{Change of prediction softmax score of MLP model  obtained from last hidden layer activation of the CNN model under BIM attack with different amount of maximum allowed perturbation ($\epsilon$) values.}
    \label{fig:dist_list_bim}
\end{figure}

In Figure \ref{fig:dist_list_bim}, we show the efficacy of our proposed method. The y-axis shows the softmax prediction scores of the MLP model for the predicted class of the CNN model. MLP model was already trained on the last hidden layer activations of the CNN model for clean, noisy and perturbed samples. Since it learnt to map all the training samples (last hidden layer activations) into their related actual class labels accurately, even the CNN classifier is fooled and predicts a wrong class for a deliberately perturbed sample, the MLP model still predicts correct output given the last hidden layer activations of the CNN model for that perturbed sample. In the figures, the green regions represent the areas of correct prediction and the red regions represent the areas of wrong prediction for the CNN model. The CNN model is fooled with $eps$ value of 0.12 and starts to make the wrong prediction. However, the MLP model prediction score decreases to 0 for the predicted class of the CNN model. 
This knowledge is already thought to the MLP model during its training. Thus, it successfully distinguishes the last hidden layer activation output of the perturbed sample as it is closer to the original class data distribution in the latent space than the target class distribution. The predicted softmax score tends to rapidly decrease to zero for the wrongly predicted class of the CNN model.

\section{Results}
\label{ch:results}

\subsection{Experimental Setup}\label{sec:experimental-setup}
We trained our CNN models for the MNIST (Digit) \cite{lecun-mnisthandwrittendigit-2010},  MNIST (Fashion) \cite{xiao2017fashionmnist} and CIFAR10 \cite{cifar10} datasets, and we achieved accuracy rates of 99.10 \% , 91.52 \% and 80.79 \% respectively. The model architectures are given in Table \ref{tab:cnn_model_arch_digit} and the hyperparameters selected in Table \ref{tab:cnn_model_params}. For training a classifier using last hidden layer activations of the CNN Models, we used simple Multilayer Perceptron (MLP) models with 2-hidden layers which are detailed in Table \ref{tab:mlp_model_arch_digit}. The hyperparameters applied for these MLP models are shown in Table \ref{tab:mlp_model_params}. In addition to the clean data, the noisy and perturbed samples which are used to train the MLP models are crafted using $eps$ values of 0.2, 0.07 and 0.03 for MNIST Digit, MNIST Fashion and CIFAR datasets respectively. Finally, we used $T = 50$ as the number of MC dropout samples when quantifying uncertainty metrics.

\begin{table}[!htbp]
    \centering 
    \caption{CNN model architectures}
    \label{tab:cnn_model_arch_digit} \scriptsize
    \begin{tabular}{|c||c|c|}
        \hline
        \textbf{Dataset} & \textbf{Layer Type} &  \textbf{Layer Information}\\
        \hline \hline
        \multirow{5.5}{*}{MNIST (Digit)} & Convolution (padding:1) + ReLU & $3 \times 3 \times 10$ \\
        & Convolution (padding:1) + ReLU & $3 \times 3 \times 20$ \\
        & Dropout & p : 0.5 \\
        & Fully Connected + ReLU & $2880 \times 128$ \\
        & Dropout & p : 0.5 \\
        & Fully Connected + ReLU & $128 \times 10$ \\
        \hline \hline
        \multirow{12}{*}{MNIST (Fashion)} & Convolution (Padding = 1) + ReLU & $3 \times 3 \times 32$ \\
        & Max Pooling & $2 \times 2$ \\
        & Convolution (Padding = 1) + ReLU & $3 \times 3 \times 32$ \\
        & Max Pooling & $2 \times 2$ \\
        & Convolution (Padding = 1) + ReLU & $3 \times 3 \times 64$ \\
        & Dropout & p : 0.25 \\
        & Convolution (Padding = 1) + ReLU & $3 \times 3 \times 64$ \\
        & Dropout & p : 0.25 \\
        & Fully Connected + ReLU & $3136 \times 600$ \\
        & Dropout & p : 0.5 \\
        & Fully Connected + ReLU & $600 \times 128$ \\
        & Fully Connected + ReLU & $128 \times 10$ \\
        \hline \hline
        \multirow{14}{*}{CIFAR10} & Convolution (Padding = 1) + ReLU & $3 \times 3 \times 32$ \\
        & Convolution (Padding = 1) + ReLU & $3 \times 3 \times 64$ \\
        & Max Pooling (Stride 2) & $2 \times 2$ \\
        & Convolution (Padding = 1) + ReLU & $3 \times 3 \times 128$ \\
        & Convolution (Padding = 1) + ReLU & $3 \times 3 \times 128$ \\
        & Max Pooling (Stride 2) & $2 \times 2$ \\
        & Dropout & p : 0.5 \\     
        & Convolution (Padding = 1) + ReLU & $3 \times 3 \times 256$ \\
        & Convolution (Padding = 1) + ReLU & $3 \times 3 \times 256$ \\
        & Max Pooling (Stride 2) & $2 \times 2$ \\        
        & Fully Connected + ReLU & $4096 \times 1024$ \\
        & Dropout & p : 0.5 \\
        & Fully Connected + ReLU & $1024 \times 256$ \\
        & Dropout & p : 0.5 \\
        & Fully Connected + ReLU & $256 \times 10$ \\
        \hline
    \end{tabular}
\end{table}

\begin{table}[!htbp]
    \centering
    \caption{CNN model parameters}
    \label{tab:cnn_model_params} \scriptsize
    \begin{tabular}{c|c|c|c} 
        \hline
        \textbf{Parameter} & \textbf{MNIST (Digit)} & \textbf{MNIST (Fashion)} & \textbf{CIFAR10} \\
        \hline \hline
        Optimizer & Adam & Adam & Adam\\
        Learning rate & 0.001 & 0.001 & 0.001\\
        Batch Size & 64 & 64 & 128\\
        Dropout Ratio & 0.5 & 0.25 & 0.5\\
        Epochs & 10 & 10 & 50\\
        \hline
    \end{tabular}
\end{table}

\begin{table}[!htbp]
    \centering
    \caption{MLP model architectures}
    \label{tab:mlp_model_arch_digit} \scriptsize
    \begin{tabular}{|c||c|c|}
        \hline
        \textbf{Dataset} & \textbf{Layer Type} &  \textbf{Layer Information}\\
        \hline \hline
        \multirow{4}{*}{MNIST (Digit)} & Fully Connected + ReLU & $128 \times 512$ \\
        & Fully Connected + ReLU & $512 \times 1024$ \\
        & Fully Connected + ReLU & $1024 \times 128$ \\
        & Fully Connected & $128 \times 10$ \\
        \hline \hline
        \multirow{4}{*}{MNIST (Fashion)} & Fully Connected + ReLU & $128 \times 512$ \\
        & Fully Connected + ReLU & $512 \times 1024$ \\
        & Fully Connected + ReLU & $1024 \times 512$ \\
        & Fully Connected & $512 \times 10$ \\
        \hline \hline
        \multirow{4}{*}{CIFAR10} & Fully Connected + ReLU & $256 \times 512$ \\
        & Fully Connected + ReLU & $512 \times 1024$ \\
        & Fully Connected + ReLU & $1024 \times 512$ \\
        & Fully Connected & $512 \times 10$ \\
        \hline
    \end{tabular}
\end{table}

\begin{table}[h]
    \centering
    \caption{MLP model parameters}
    \label{tab:mlp_model_params} \scriptsize
    \begin{tabular}{c|c|c|c} 
        \hline
        \textbf{Parameter} & \textbf{MNIST (Digit)} & \textbf{MNIST (Fashion)} & \textbf{CIFAR10} \\
        \hline \hline
        Optimizer & Adam & Adam & Adam\\
        Learning rate & 0.001 & 0.001 & 0.001\\
        Batch Size & 128 & 128 & 128\\
        Epochs & 50 & 50 & 150\\
        \hline
    \end{tabular}
\end{table}

\subsection{Experimental Results}\label{sec:experimental-results}

\begin{figure}[htbp!]
     \centering
     \begin{subfigure}[b]{0.32\linewidth}
         \centering
         \includegraphics[width=\linewidth]{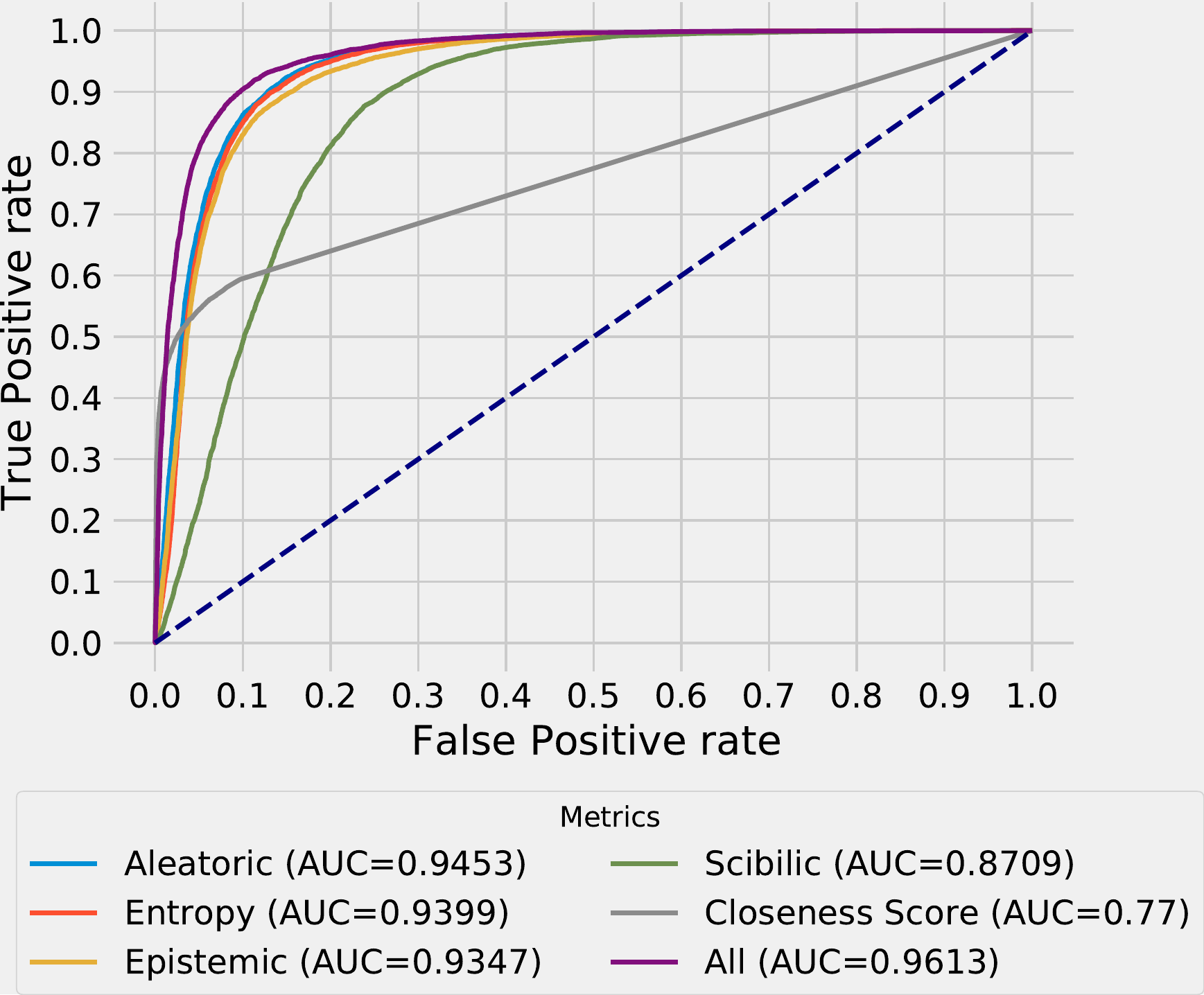}
         \caption{BIM}
         \label{fig:ROC_MNIST_BIM_012}
     \end{subfigure}
     \begin{subfigure}[b]{0.32\linewidth}
         \centering
         \includegraphics[width=\linewidth]{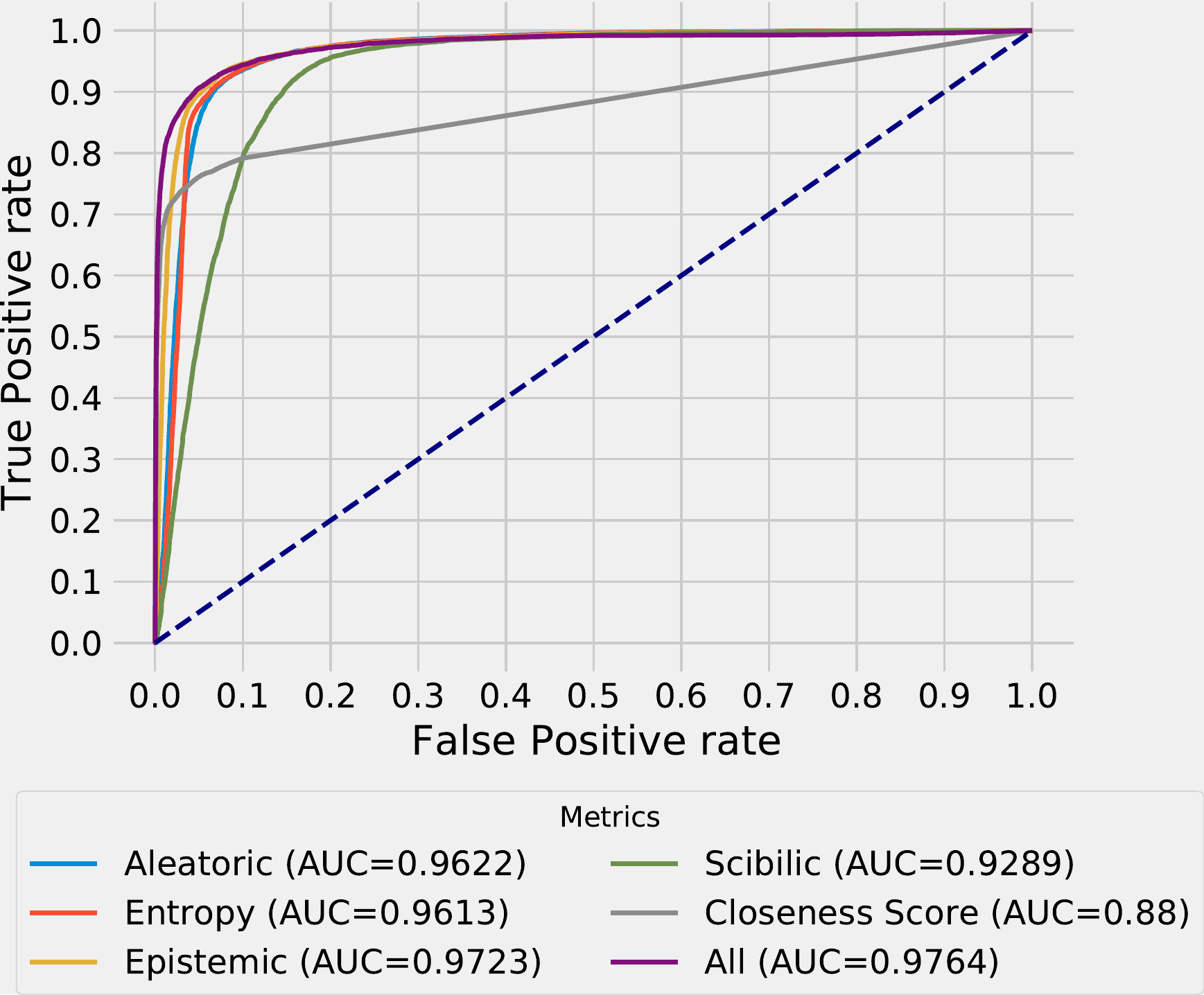}
         \caption{CW}
         \label{fig:ROC_MNIST_CW_012}
     \end{subfigure}
     \begin{subfigure}[b]{0.32\linewidth}
         \centering
         \includegraphics[width=\linewidth]{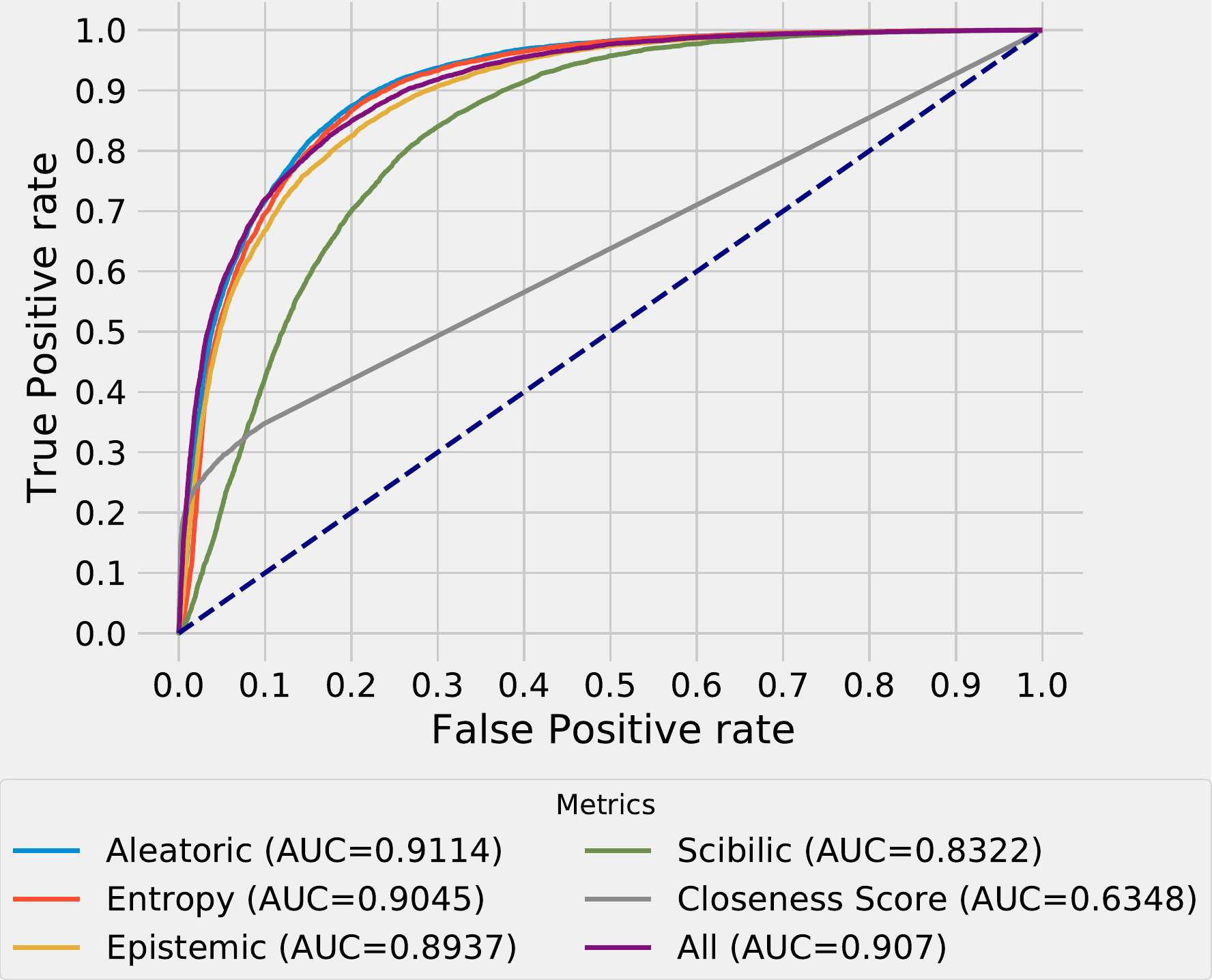}
         \caption{DeepFool}
         \label{fig:ROC_MNIST_Deepfool_012}
     \end{subfigure}
     \begin{subfigure}[b]{0.32\linewidth}
         \centering
         \includegraphics[width=\linewidth]{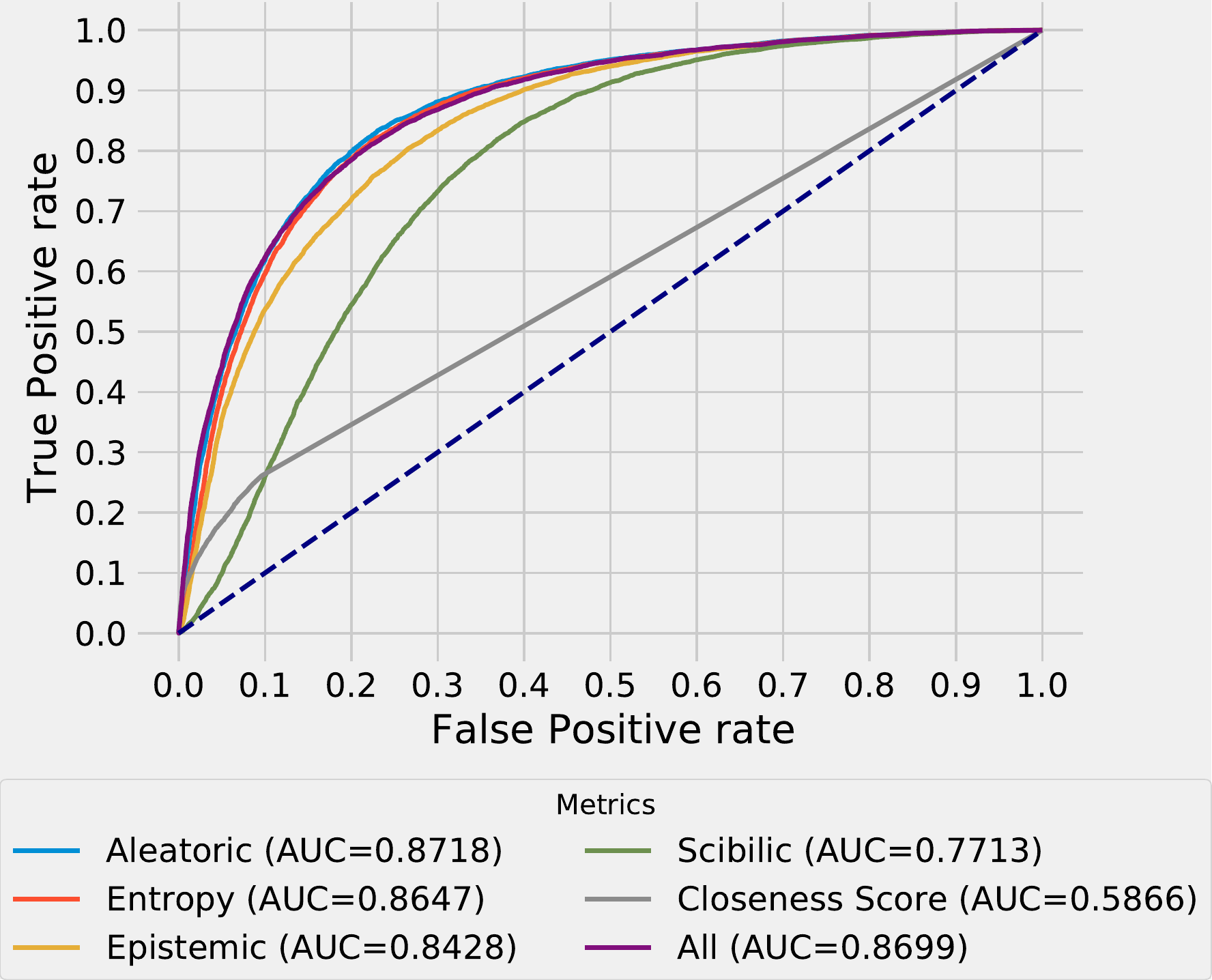}
         \caption{FGSM}
         \label{fig:ROC_MNIST_FGSM_012}
     \end{subfigure}
     \begin{subfigure}[b]{0.32\linewidth}
         \centering
         \includegraphics[width=\linewidth]{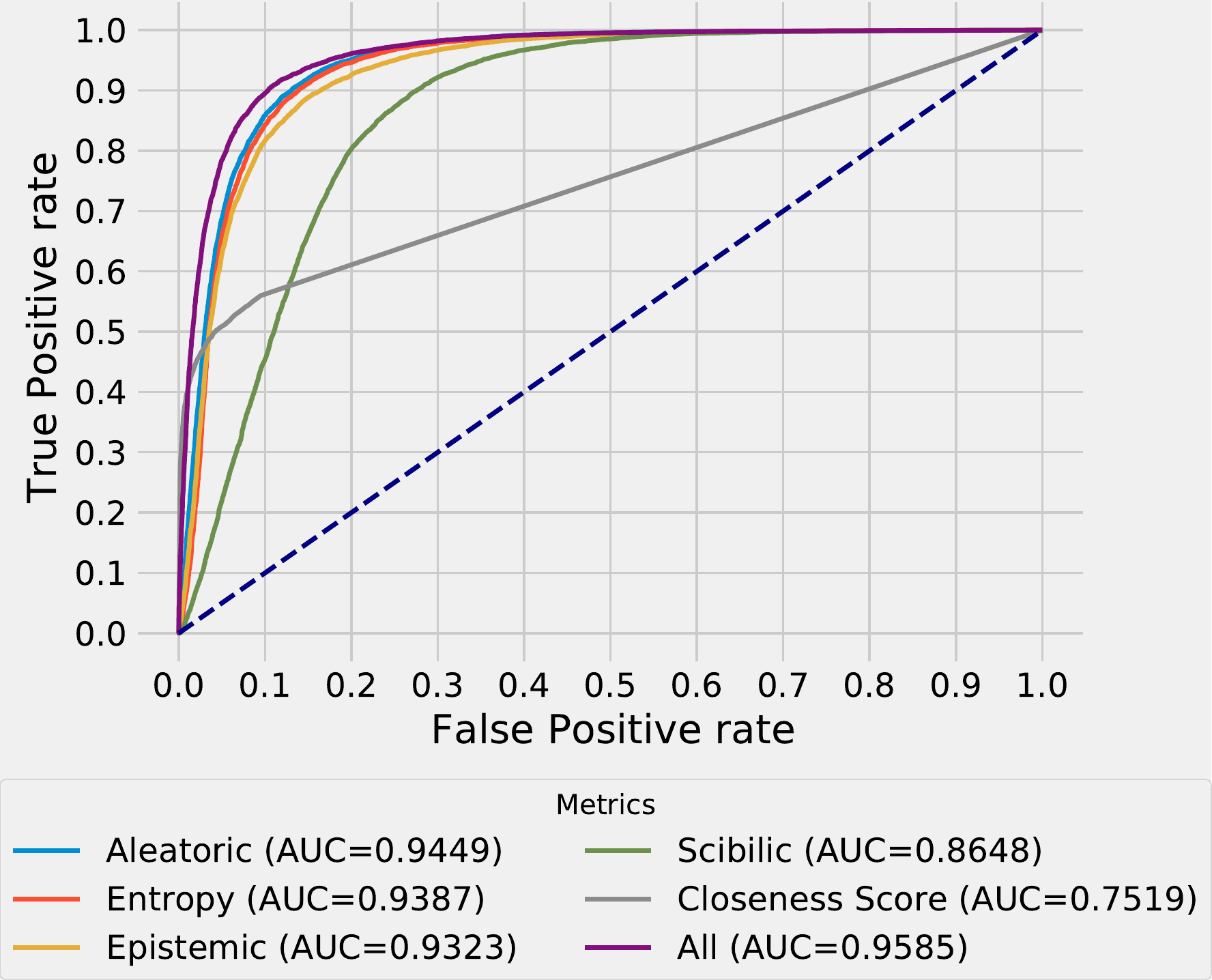}
         \caption{PGD}
         \label{fig:ROC_MNIST_PGD_012}
     \end{subfigure}
        \caption{MNIST - $\epsilon=0.12$}
        \label{fig:MNIST_ROC_012}
\label{fig:ROC_012}
\end{figure}

\begin{figure}[htbp!]
     \centering
     \begin{subfigure}[b]{0.32\linewidth}
         \centering
         \includegraphics[width=\linewidth]{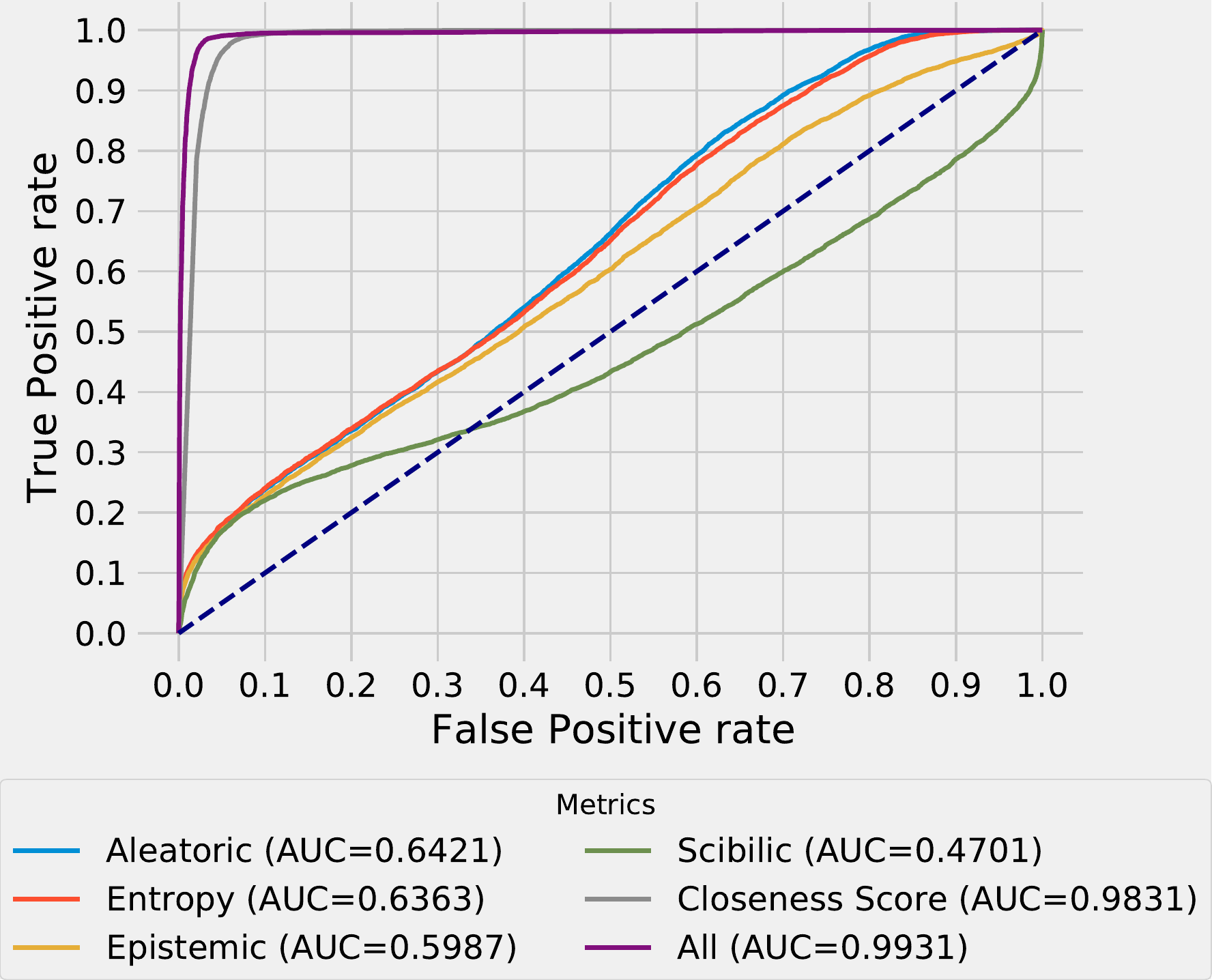}
         \caption{BIM}
         \label{fig:ROC_MNIST_BIM_030}
     \end{subfigure}
     \begin{subfigure}[b]{0.32\linewidth}
         \centering
         \includegraphics[width=\linewidth]{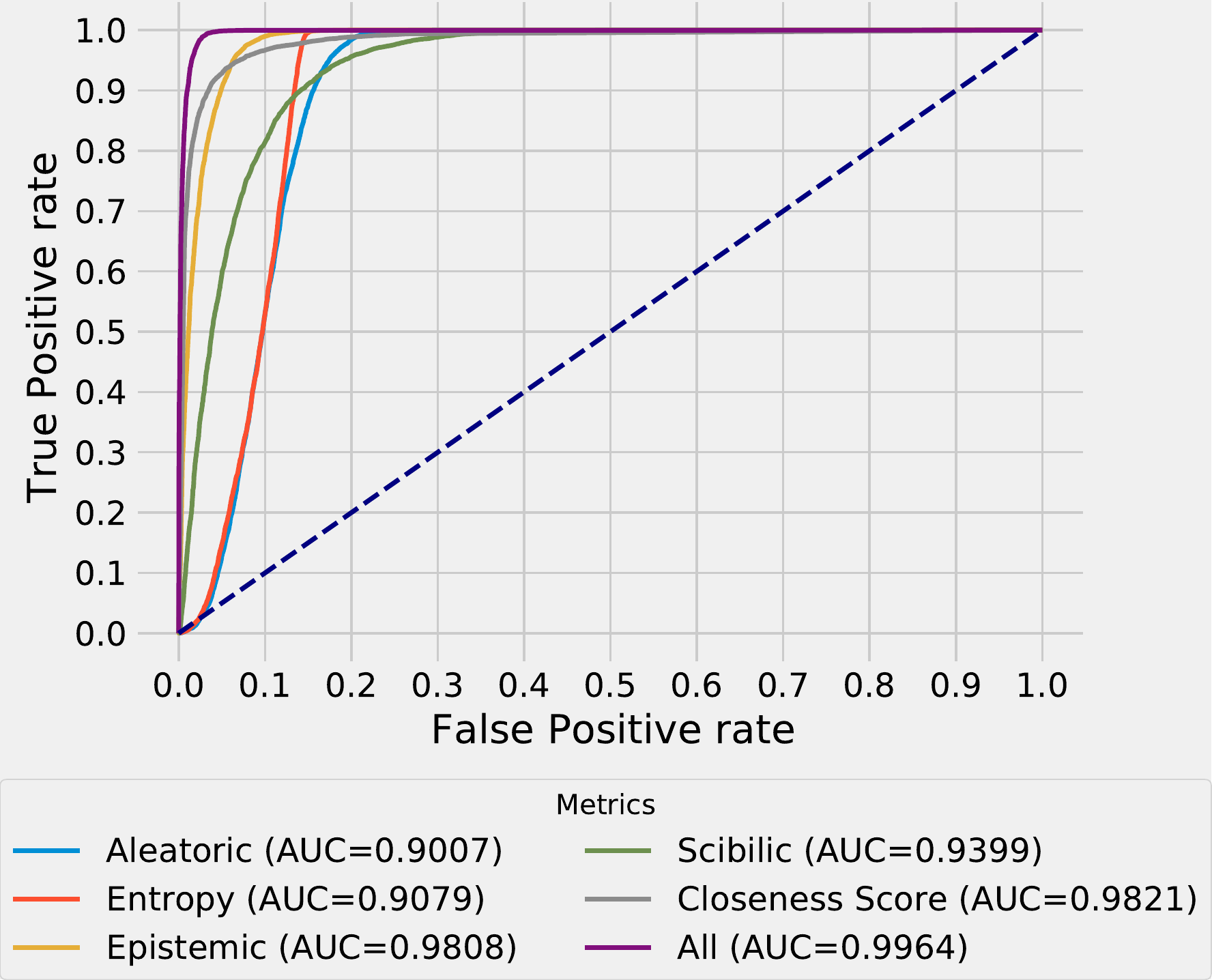}
         \caption{CW}
         \label{fig:ROC_MNIST_CW_030}
     \end{subfigure}
     \begin{subfigure}[b]{0.32\linewidth}
         \centering
         \includegraphics[width=\linewidth]{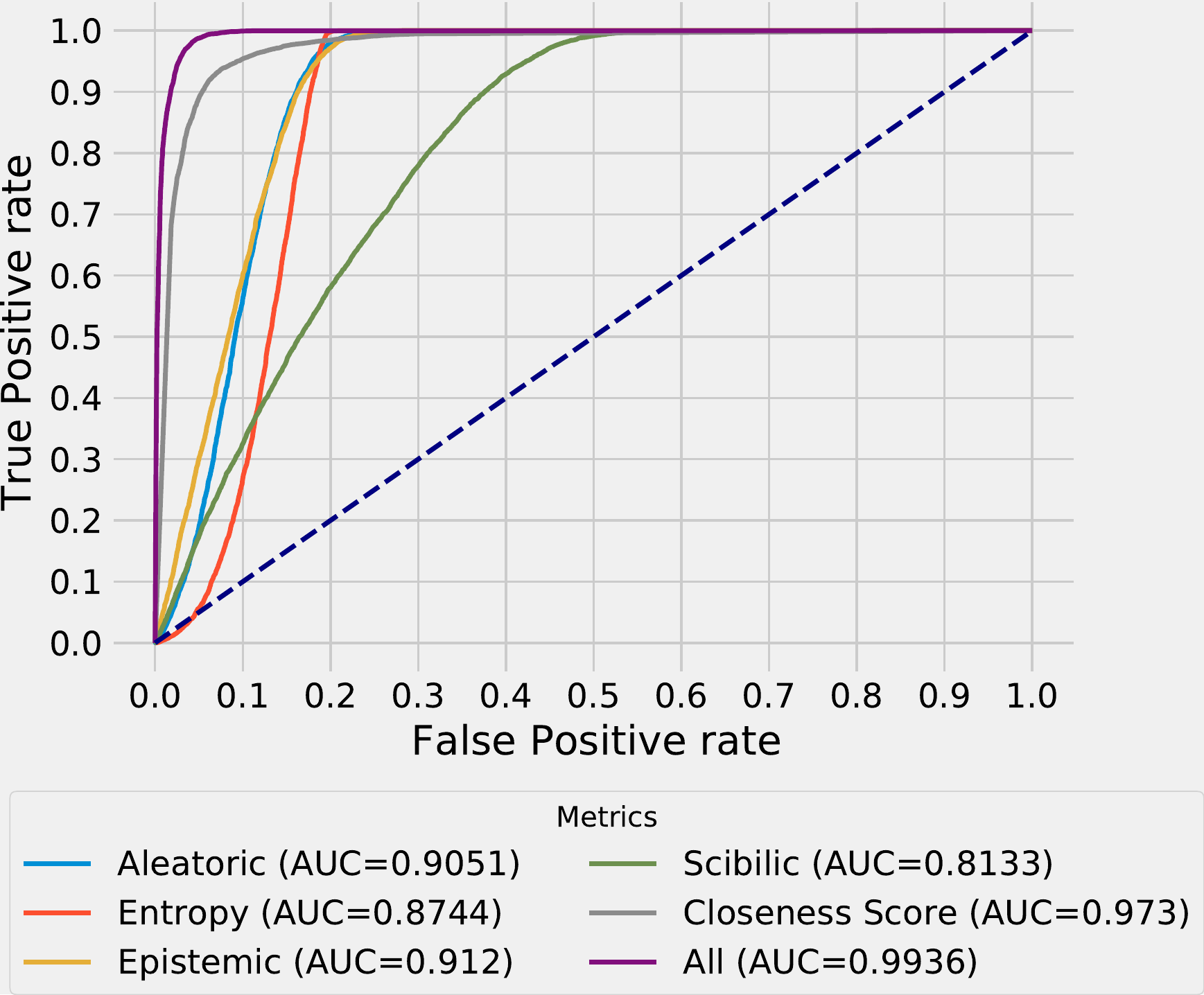}
         \caption{DeepFool}
         \label{fig:ROC_MNIST_Deepfool_030}
     \end{subfigure}
     \begin{subfigure}[b]{0.32\linewidth}
         \centering
         \includegraphics[width=\linewidth]{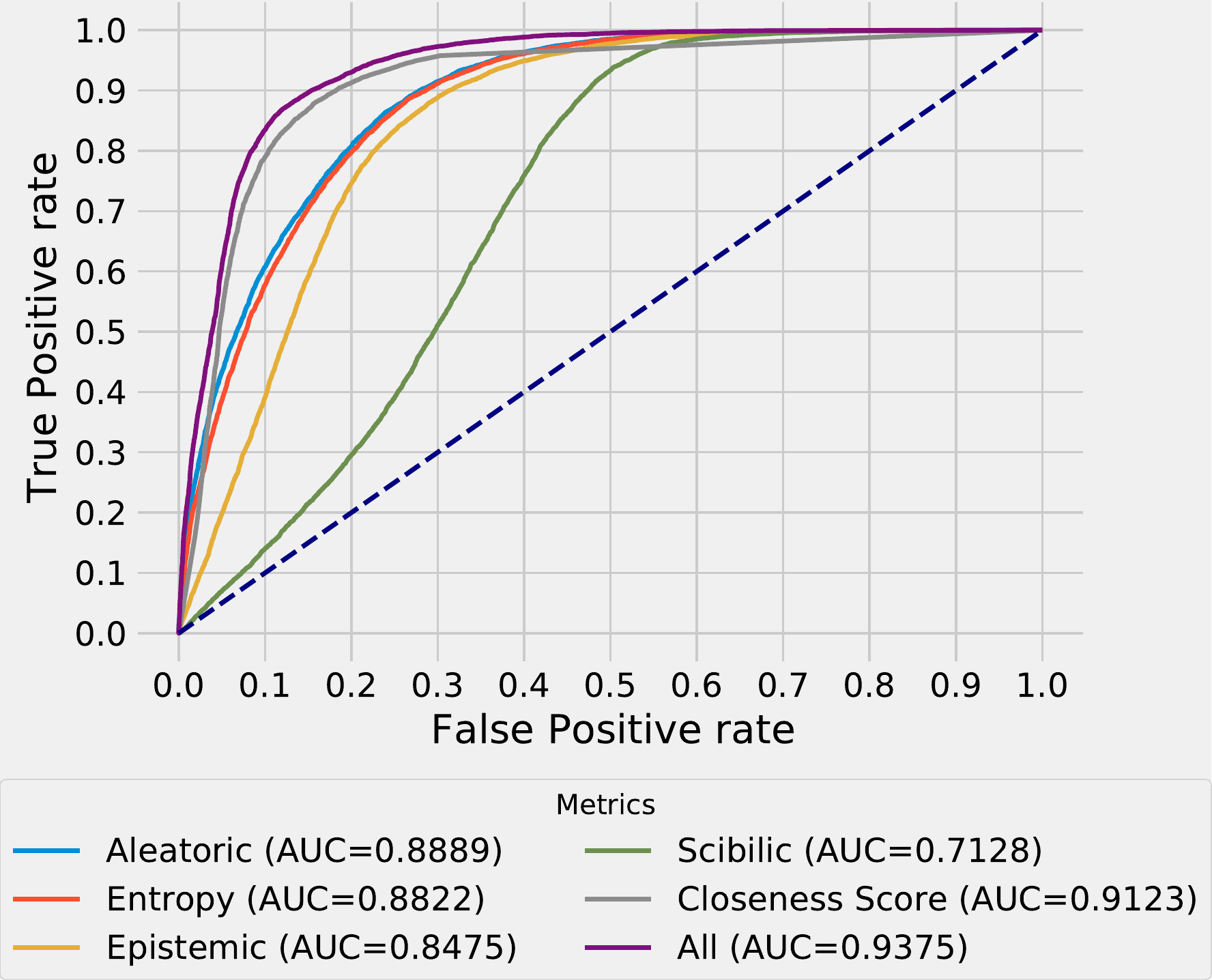}
         \caption{FGSM}
         \label{fig:ROC_MNIST_FGSM_030}
     \end{subfigure}
     \begin{subfigure}[b]{0.32\linewidth}
         \centering
         \includegraphics[width=\linewidth]{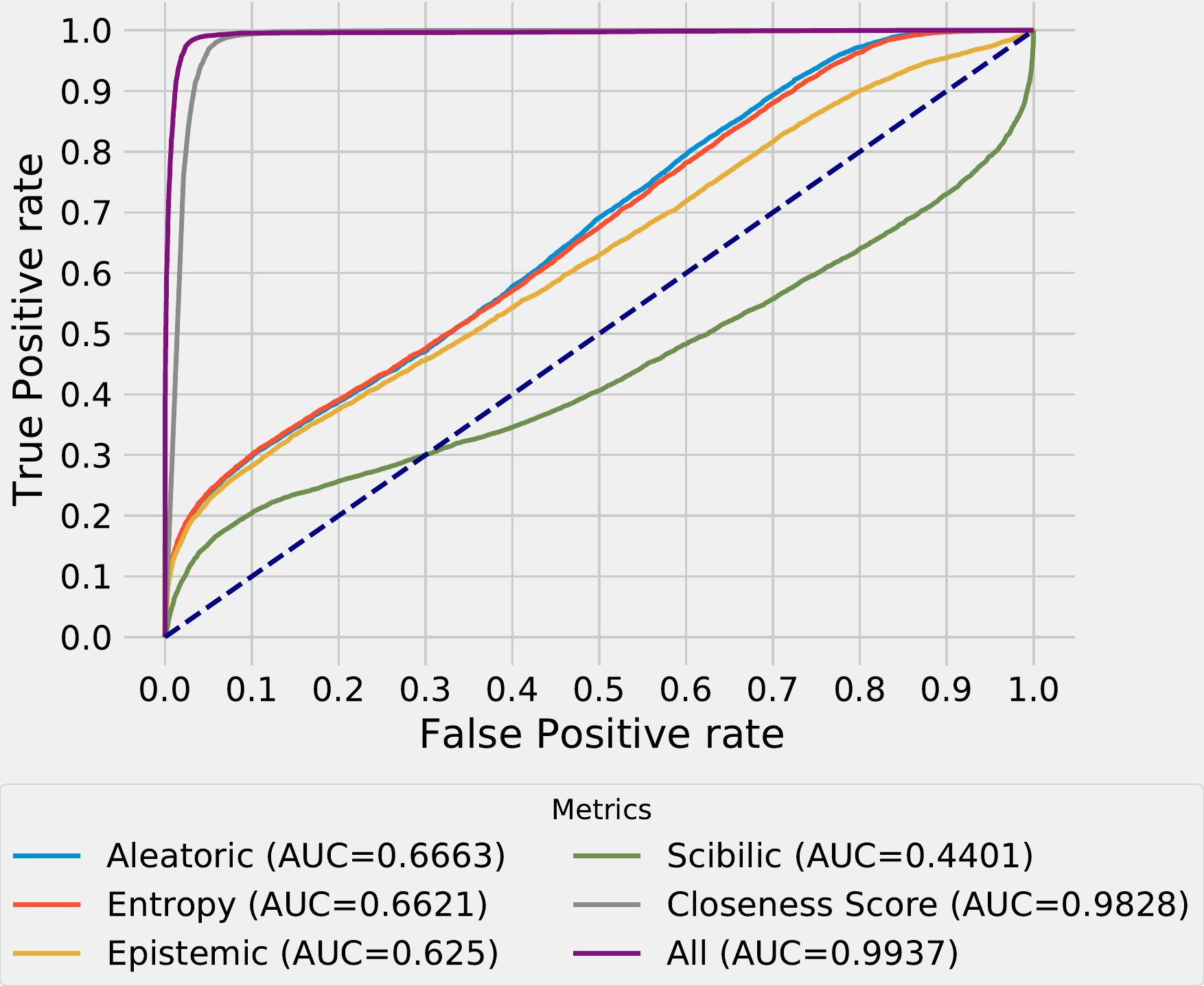}
         \caption{PGD}
         \label{fig:ROC_MNIST_PGD_030}
     \end{subfigure}
        \caption{MNIST - $\epsilon=0.30$}
        \label{fig:MNIST_ROC_030}
\label{fig:ROC_030}
\end{figure}

To evaluate the performance of different metrics for adversarial detection, we have implemented each of the 5 attacks (FGSM, BIM, PGD, CW and Deepfool) with different allowed perturbation amounts ($\epsilon$) under $l_{inf}$ norm on MNIST (Digit), MNIST (Fashion) and CIFAR10 test data. Just for CW attack, we used the $l_{2}$ norm equivalent of the applied perturbation by using the formula $l_{2}  = l_{inf}  \times \sqrt{n} \times \sqrt{2}/\sqrt{\pi e}$, where $n$ is the input sample dimension. For the implementations of these attacks, we used a Python toolbox called Foolbox \cite{rauber2018foolbox} and implement the attacks in their default settings. To be consistent with Feinman et al. \cite{feinman2017detecting}, we only perturbed those test samples which were correctly classified by our models in their original states. Because an adversary would have no reason to perturb samples that are already misclassified. We have also included normal and noisy counterparts in the pool for each adversarial sample as a benchmark. We craft noisy samples by applying Gaussian noise to each pixel with a scale similar to the adversarial samples. Then, all these normal, noisy and perturbed samples are used to train a Logistic Regression (LR) model to test the performance of our adversarial classifier. Adversarial samples are labeled as 1, representing the positive class, whereas normal and noisy samples are labeled as 0, representing the negative class. Five features that are computed for each sample in the pool before LR training are Epistemic Uncertainty, Aleatoric Uncertainty, Scibilic Uncertainty, Entropy and Closeness Score for predicted class. For MNIST (Digit) dataset, we showed ROC-AUC scores of our adversarial classifier as in Figures \ref{fig:ROC_012} and \ref{fig:ROC_030}. And the detailed results of our experiments are available in Tables \ref{tab:my_label_Digit} , \ref{tab:my_label_Fashion} and \ref{tab:my_label_CIFAR}. We aimed to evaluate our metrics' quality under both medium and high level of adversarial threat by using different allowed perturbation amounts. We achieved almost perfect detection scores under the high level of allowed perturbation (epsilon). And despite the risk of lowering attack success chance, if the intruder opts to choose a lower epsilon value for the attack, we again achieve a very high degree of performance.

\begin{table}[htbp!]
\caption{Digit MNIST - Roc-Auc Scores of different metrics under various attack types and epsilon values}
    \label{tab:my_label_Digit}
    \centering \footnotesize
    \begin{tabular}{|c|c|c|c|c|c|c||c|c|c|c|c|c|} \hline
    & \multicolumn{6}{c||}{eps = 0.12} & \multicolumn{6}{c|}{eps = 0.30} \\ \cline{2-13}
          & \textbf{Epis.} & \textbf{Alea.} & \textbf{Scibilic} & \textbf{Ent.} & \textbf{Dist.} & \textbf{All} & \textbf{Epis.} & \textbf{Alea.} & \textbf{Scibilic} & \textbf{Ent.} & \textbf{Dist.} & \textbf{All} \\ \hline
\textbf{FGSM} & 0.84 & 0.87 & 0.77 & 0.86 & 0.59 & 0.87 & 0.85 & 0.89 & 0.71 & 0.88 & 0.91 & 0.94 \\
\textbf{BIM} & 0.93 & 0.95 & 0.87 & 0.94 & 0.77 & 0.96 & 0.60 & 0.64 & 0.47 & 0.64 & 0.98 & 0.99 \\
\textbf{PGD} & 0.93 & 0.94 & 0.86 & 0.94 & 0.75 & 0.96 & 0.63 & 0.67 & 0.44 & 0.66 & 0.98 & 0.99 \\
\textbf{Deepfool} & 0.89 & 0.91 & 0.83 & 0.90 & 0.63 & 0.91 & 0.91 & 0.91 & 0.81 & 0.87 & 0.97 & 0.99 \\
\textbf{CW} & 0.97 & 0.96 & 0.93 & 0.96 & 0.88 & 0.98 & 0.98 & 0.90 & 0.94 & 0.91 & 0.98 & 1.00 \\ \hline
    \end{tabular}
\end{table}

\begin{table}[htbp!]
\caption{Fashion MNIST - Roc-Auc Scores of different metrics under various attack types and epsilon values}
    \label{tab:my_label_Fashion}
    \centering \footnotesize
    \begin{tabular}{|c|c|c|c|c|c|c||c|c|c|c|c|c|} \hline
    & \multicolumn{6}{c||}{eps = 0.03} & \multicolumn{6}{c|}{eps = 0.12} \\ \cline{2-13}
          & \textbf{Epis.} & \textbf{Alea.} & \textbf{Scibilic} & \textbf{Ent.} & \textbf{Dist.} & \textbf{All} & \textbf{Epis.} & \textbf{Alea.} & \textbf{Scibilic} & \textbf{Ent.} & \textbf{Dist.} & \textbf{All} \\ \hline
\textbf{FGSM} & 0.77 & 0.76 & 0.77 & 0.76 & 0.61 & 0.78 & 0.80 & 0.77 & 0.79 & 0.79 & 0.76 & 0.86 \\
\textbf{BIM} & 0.77 & 0.72 & 0.78 & 0.74 & 0.74 & 0.85 & 0.69 & 0.72 & 0.33 & 0.72 & 0.99 & 0.99 \\
\textbf{PGD} & 0.78 & 0.74 & 0.79 & 0.75 & 0.71 & 0.84 & 0.71 & 0.73 & 0.69 & 0.73 & 0.99 & 0.99 \\
\textbf{Deepfool} & 0.89 & 0.85 & 0.88 & 0.86 & 0.76 & 0.90 & 0.95 & 0.89 & 0.91 & 0.90 & 0.93 & 0.98 \\
\textbf{CW} & 0.89 & 0.84 & 0.89 & 0.85 & 0.79 & 0.91 & 0.96 & 0.87 & 0.93 & 0.89 & 0.94 & 0.98 \\ \hline
    \end{tabular}
\end{table}

\begin{table}[htbp!]
\caption{CIFAR10 - Roc-Auc Scores of different metrics under various attack types and epsilon values}
    \label{tab:my_label_CIFAR}
    \centering \footnotesize
    \begin{tabular}{|c|c|c|c|c|c|c||c|c|c|c|c|c|} \hline
    & \multicolumn{6}{c||}{eps = 0.02} & \multicolumn{6}{c|}{eps = 0.04} \\ \cline{2-13}
          & \textbf{Epis.} & \textbf{Alea.} & \textbf{Scibilic} & \textbf{Ent.} & \textbf{Dist.} & \textbf{All} & \textbf{Epis.} & \textbf{Alea.} & \textbf{Scibilic} & \textbf{Ent.} & \textbf{Dist.} & \textbf{All} \\ \hline
\textbf{FGSM} & 0.70 & 0.69 & 0.68 & 0.69 & 0.53 & 0.71 & 0.71 & 0.69 & 0.68 & 0.70 & 0.55 & 0.72 \\
\textbf{BIM} & 0.82 & 0.84 & 0.83 & 0.84 & 0.89 & 0.92 & 0.89 & 0.94 & 0.95 & 0.95 & 0.96 & 0.99 \\
\textbf{PGD} & 0.77 & 0.79 & 0.77 & 0.79 & 0.84 & 0.87 & 0.89 & 0.94 & 0.93 & 0.94 & 0.95 & 0.98 \\
\textbf{Deepfool} & 0.94 & 0.88 & 0.83 & 0.89 & 0.77 & 0.96 & 0.93 & 0.87 & 0.83 & 0.88 & 0.77 & 0.96 \\
\textbf{CW} & 0.94 & 0.87 & 0.85 & 0.88 & 0.79 & 0.97 & 0.93 & 0.85 & 0.84 & 0.86 & 0.80 & 0.96 \\ \hline
    \end{tabular}
\end{table}

\subsection{Further Results and Discussion}\label{sec:discuss}

\begin{figure}[!htbp]
    \centering
    \includegraphics[width=0.6\linewidth]{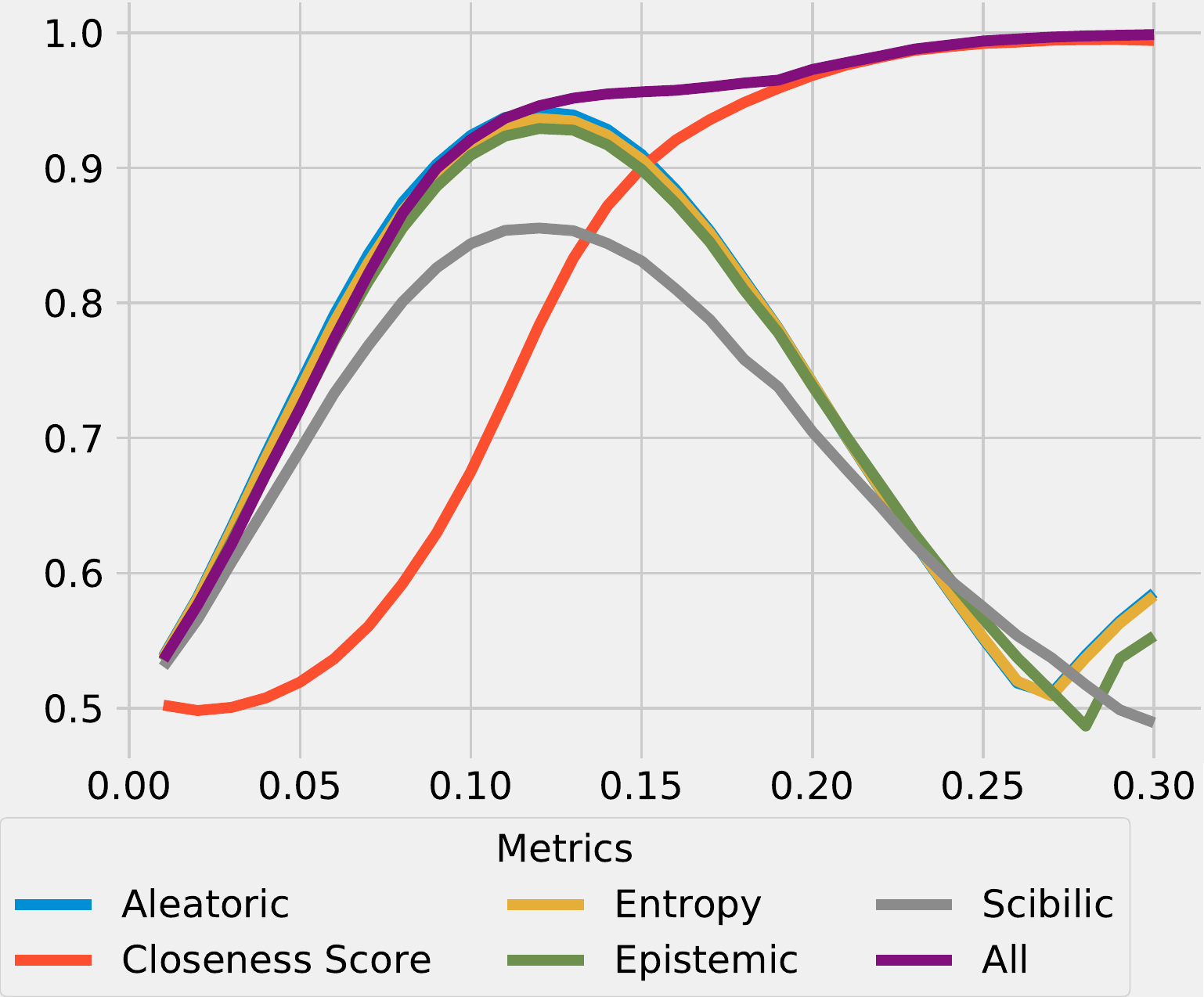}
    \caption{ROC-AUC scores of different metrics under an attack with varying level of allowed perturbation amounts $\epsilon$}
    \label{BIM_metrics}
\end{figure}

We have finally made an in-depth series of experiments to see the performance of each of the metrics under the application of an adversarial attack (BIM) with different level of perturbation amounts. Figure \ref{BIM_metrics} summarizes the results of our experiments. Results show that when the perturbation amount is low or moderate, the contribution of uncertainty metrics to adversarial detection performance is high. However, when we apply the attack with a high level of perturbation, our closeness metric takes the lead and plays the key role. The closeness metric performs poorly under an attack with a low perturbation amount because for those cases, the attack success rates are actually not so high, and the attack barely succeeds in fooling the CNN Model.  Therefore, the predictions of the CNN and MLP models are mostly the same, and the softmax output score of the MLP model for the predicted class can not act as a successful separator for clean and perturbed samples for the LR classifier. Ultimately, the combined usage of all the metrics is observed to be the best choice for securing the model prediction performance.

\section{Conclusion}
\label{ch:conclusion}

In this study, we analyzed the usage of different metrics for adversarial sample detection and showed that moment-based predictive uncertainty estimates together with the closeness score for predicted class obtained from the last hidden layer activation's can be effectively used as a tool for a successful defense mechanism against adversarial attacks. We have tested and verified our approach’s effectiveness in three different benchmark datasets, which are heavily used by the adversarial research community. The results of our comprehensive experiments show that our proposed method achieves impressive ROC-AUC scores on a range of datasets and generalize well across different attack types. In the final analysis, we have demonstrated the contribution of different metrics to adversarial sample detection under an attack with variable strength levels.

In the present study, we have only concentrated on the image domain and used CNN architectures. Nevertheless, we question whether different uncertainty metrics for adversarial sample detection are applicable to other domains like text in which different architectures are used. Hence, we plan to utilize and test the effectiveness of our metrics on different DNN architectures used in other domains as well. Besides, one other potential direction for us is to examine the usage of aleatoric and scibilic uncertainty to develop new adversarial attack ideas and achieve the goal of building more robust models.

\bibliographystyle{elsarticle-num} 
\bibliography{references}

\end{document}